\documentclass[12pt,a4paper]{article}
\usepackage{tikz}
\usepackage{graphicx}
\usepackage{shortcuts}
\usepackage{hyperref}

\usepackage[authoryear, round]{natbib}
\bibpunct[, ]{(}{)}{,}{a}{}{,}%

\usepackage[utf8]{inputenc}
\usepackage[T1]{fontenc}
\usepackage{url}
\usepackage{booktabs}
\usepackage{amsfonts}
\usepackage{nicefrac}
\usepackage{lmodern}
\usepackage{xcolor}
\usepackage{amsmath,amsthm,amssymb}
\newtheorem{lemma}{Lemma}
\newtheorem{theorem}{Theorem}

\newtheorem{assumption}{Assumption}
\newtheorem{definition}{Definition}
\newtheorem{remark}{Remark}
\newtheorem{example}{Example}

\usepackage[margin=2.4cm]{geometry}
\usepackage{setspace}
\AtBeginDocument{\doublespacing}
\usepackage{mathtools}
\usepackage{enumitem}
\usepackage{authblk}
\usepackage{tikz}
\usepackage{amsmath,amssymb}
\usepackage{algorithm}
\usepackage{algorithmic}
\usepackage{caption}

\usetikzlibrary{arrows.meta,shapes.geometric}

\usepackage[compact]{titlesec}

\title{
Agent Evaluation Without Ground Truth: 
\\  A Null Proportion Perspective
}
\author[1]{Brian M. Cho}
\author[2]{Michael Lindon}
\author[1,2]{Nathan Kallus}
\affil[1]{Cornell Tech}
\affil[2]{Netflix}
\date{}

\begin{document}
\maketitle
\setlength{\abovedisplayskip}{8pt plus 2pt minus 3pt}
\setlength{\belowdisplayskip}{8pt plus 2pt minus 3pt}
\setlength{\abovedisplayshortskip}{3pt plus 2pt minus 1pt}
\setlength{\belowdisplayshortskip}{5pt plus 2pt minus 2pt}

\begin{abstract}


The correctness of actions taken by agents is typically evaluated against known, externally verified ground truth labels, such as whether a proof is correct or whether code compiles and passes unit tests. However, as coding and science agents are entrusted with more end-to-end autonomy -- proposing interventions, running experiments, and deciding what to deploy and investigate next -- whether an agent's actions genuinely improve an outcome must be inferred from noisy, often inconclusive experiments.
To evaluate agents in this setting, we define agent quality as a weighted sum of correct decisions along its trajectory, using sequences of $e$-values as statistical evidence regarding correctness in place of ground-truth labels. Our framework accommodates arbitrarily dependent decisions, data-dependent trajectory termination, dynamically generated hypotheses, and quality measures defined after the trajectory concludes. We construct upper and lower bound estimators and confidence sets for agent quality with statistical guarantees that hold uniformly over agent trajectory distributions and establish instance-wise sharpness of our estimator class. We demonstrate our methods with a case study comparing the success of Claude Opus and Claude Haiku in improving the runtime of an open-source Python package.
\end{abstract}

\section{Introduction}

In recent years, agentic AI has proliferated across real-world systems. Coding agents routinely propose changes to production software and infrastructure that are then deployed to its user population \citep{alshahwan2024automatedunittestimprovement, novikov2025alphaevolvecodingagentscientific}. Multi-agent systems are used to generate scientific hypotheses \citep{Gottweis_2026, xie2026scientificobservationsmechanismsbenchmarking}, which are then tested in real-world experiments. As agents are given authority to make autonomous decisions, a fundamental task is reliable evaluation for the quality of an agent's decisions.

While benchmarks remains the de-facto choice for measuring agent quality, these methods fall short when assessing the decision made by an agent during deployment. Consider an autonomous coding agent that updates the user interface, with the goal of increasing average user engagement. A change that increases user engagement from the baseline is considered correct. In traditional software-based benchmarks, such as SWE-Bench \citep{jimenez2024swebenchlanguagemodelsresolve}, each decision can be labeled as correct or incorrect with unit tests or reference answers. For our coding agent, however, whether its changes actually improved average user engagement is unknown: it corresponds to a statement about the unknown distribution of the system. While data may be collected to test this claim, the expected increase in user engagement over the entire user population is \textit{never directly observed}. Our work focuses this setting, where correctness is never observed, and only noisy measurements are available to serve as evidence for and against the correctness of an agent's decision. 


To deal with unknown decision quality, popular approaches are the use of estimated correctness labels, from either statistical methods or another agent. Popular heuristics include the LLM-as-judge framework, where a separate model is used a judge for the agent's decisions  \citep{zheng2023judging}. Alternative frameworks use a small set of decisions with labeled correctness to construct estimated labels \citep{wang2020methods, angelopoulos2023ppi}. In our setting, however, neither approach provides reliable evaluation with statistical guarantees. Imputed labels by LLMs are prone to substantial bias and invalid confidence intervals, even with relatively high accuracy for estimated labels \citep{egami2023using}. While statistical frameworks such as prediction-powered inference construct unbiased correctness estimates \citep{boyeau2025autoeval}, these methods require that at least a small set of true labels are observed, failing to capture our setting with no observed labels.

In this work, we cast the problem of agentic evaluation as a multiple hypothesis testing problem, where each decision is associated with the hypotheses regarding the correctness of that decision. For example, in a software deployment setting, each decision is associated with a pair of hypotheses claiming that the change increased or decreased user engagement, with a \textit{correct} change being defined as one that increases expected engagement. Agent quality is defined as a weighted sum of correct
decisions, such as the proportion of the agent's changes that improved
engagement. Our framework corresponds to evaluating an agent's decisions as a weighted sum of null hypothesis counts, building upon null proportion estimation \citep{storey2002direct, meinshausen2006estimating, efron2008microarrays}.

Distinct from existing works, we develop methods tailored to three key features of agent applications in practice. First, agents act adaptively: each decision is chosen in response of the outcomes of earlier ones and the system shifts in response. Consequently, decisions and the evidence on their quality may be arbitrarily dependent. Second, agent applications are often monitored continuously, and may be terminated at any data-dependent time. For example, when data collected about the agent indicates harmful behavior, it is typically terminated in response. Third, quality may be defined only after the trajectory concludes; the proportion of improving changes, for example, depends on whether the agent chose to adopt a change and how many total changes were proposed.
Classical null estimation settings assume a fixed collection of hypotheses with $p$-values available at a fixed time. In agent applications, however, the number of changes may be random, and classical $p$-values are often invalid under data-driven stopping.

Our primitive for evidence is the $e$-value \citep{vovk2021evalues,
grunwald2024safe, ramdas2023gametheoretic}. For each decision we use two sequences of $e$-values that test the correctness of the agent's decision during deployment. Unlike $p$-values, $e$-values remain valid under optional stopping and can be combined under arbitrary dependence, enabling the flexibility required for agentic evaluation. They are also readily available: anytime-valid tests based on $e$-values are already used to gate software releases and to
monitor experiments in industrial applications \citep{lindon2022rapid,
johari2022always}.

Our contributions are as follows.
\begin{enumerate}
  \item \textbf{Framework.} We formalize agent quality as a data-dependent weighted sum of unobserved correctness indicators along an adaptively generated trajectory. Our model allows for the number of decisions, trajectory horizons, and definition of agent quality to be random and data-dependent.
  
  \item \textbf{Estimation.} We construct $e$-value-based upper and lower bound estimators that are uniformly valid over deployment trajectories. Our estimator class contains estimators that are nearly sharp estimators for any instance. In a simplified setting, we show our method dominates Storey's estimator \citep{storey2002direct}.

  \item \textbf{Inference.} We construct the confidence set for agent quality that provide uniform coverage over trajectory deployments. We relate our class with $e$-value merging and discuss the effect of random hypotheses on confidence set construction.

  \item \textbf{Real World Studies.} To test our class of estimators and confidence sets, we construct estimators and confidence intervals with predictable tuning heuristics based on trajectory simulation. We demonstrate the practical application of our methods with case studies comparing agent performance for code improvement on the open-source \texttt{NetworkX} Python package \citep{SciPyProceedings_11}.
\end{enumerate}

Section~\ref{sec:setup} formalizes agentic evaluation and the identification problem. Section~\ref{sec:estimator} and Section~\ref{sec:inference} develops the estimators and the confidence intervals respectively. Section~\ref{sec:experiments} presents our real-world case study. Further discussion on related works, proofs, tuning procedures, and experiment settings are provided in the supplement.

\section{Setup and Estimand}\label{sec:setup}

The agent collects information from a system sequentially over discrete time $t \in \NN$. It can make decisions at any point based on information observed so far.
The trajectory concludes at potentially data-driven time $N$, with $M$ decisions made by the end of the trajectory.

We seek to evaluate the quality of the agent's sequence of decisions along this trajectory. As the agent makes decisions, it induces a random sequence of binary correctness labels $(\theta_1, \dots \theta_M)$, where $\theta_i =1$ denotes that the decision was correct. We define agent quality $\theta_*$ as a weighted sum of the correctness of decisions over its trajectory,
\begin{equation}
    \theta_*=\sum_{i=1}^M c_i\theta_i, 
\end{equation}
where $c_i$ corresponds to data-driven weights that may depend on our entire observed trajectory. Simple choices of $c_i$, such as $c_i = 1$ or $c_i = 1/M$, define $\theta_*$ as the number or proportion of correct changes respectively. We provide a real-world example of $\theta_*$ below.



\begin{example}[Coding agent with live experiments]\label{ex:code_agent}
    Consider a coding agent that pushes software changes to a real-world system. Let $P_i$ denote distributions over outcome metrics $(X_{\text{goal}},X_{\text{guard}})$, such as user engagement and system crashes, after the agent's $i$-th proposed code change. Let $c_i = 1$ if the agent chooses to adopt change $i$ (such as based on an experiment yielding samples from $P_i$), and $c_i = 0$ if the change is reverted, with $j_i=\sup\{j\in\mathbb N_0:j<i,c_j=1\}$. Define correctness as whether the change improves the average goal metric by at least 10\% without deteriorating the guard metric: 
    $\theta_i = \bm 1\left\{ \EE_{P_i}[X_{\text{goal}}] / \EE_{P_{j_i}}[X_{\text{goal}}]\geq1.1 \wedge \EE_{P_i}[X_{\text{guard}}] \leq \EE_{P_{j_i}}[X_{\text{guard}}]\right\}.$
    Then, $\theta_*$ denotes the number of safe improvements generated by the agent. Dividing $c_i$ by the number of adopted changes would instead target a fraction of adopted changes that are safe improvements.
\end{example}
A similar example is the \textit{science agent} that can make hypotheses, explore them via data collection, and declare discoveries.
Both the hypotheses and how the agent investigates them can evolve fully dynamically over the trajectory.
Then, $\theta_*$ can evaluate the number or fraction of true hypotheses or true discoveries.

\subsection{Data Generating Process}

As the defining feature of real-world evaluation, the correctness labels $(\theta_1, \dots, \theta_M)$ and target $\theta_*$ are \textit{unobserved}. To assess the correctness of each decision, we consider the setting where for each $i\in[M]$, we observe a random sequence of nonnegative outputs $\mathbf{e}_i=(e_{i,t},\bar e_{i,t})_{t\in\mathbb N}$, with binary observation indicators $\mathbf{R}_i=(R_{i,t})_{t\in\mathbb N}$. The term $R_{i,t}=1$ indicates that a pair of nonnegative random variables $(e_{i,t}, \bar{e}_{i,t})$ for change $i$ is observed at time $t$. For any $(i,t)$ with $i>M$, $t>N$, or $R_{i,t}= 0$, we define $(e_{i,t},\bar e_{i,t})=(1,1)$, and set $R_{i,t} = 0$ whenever $i > M$ or $t > N$. We define the filtration over observed information as
\begin{equation}
    \Fcal_t=\sigma\big(\left\{R_{i,j}e_{i,j}, R_{i,j}\bar e_{i,j}\right\}_{(i,j) \in \NN \times [t]}, \left\{R_{i,j}\right\}_{(i,j) \in \NN \times [t+1]}, \left\{\mathbf{1}[j \leq N]\right\}_{ j \in [t+1]}\big),
\end{equation}
where $\Fcal_t$ denotes sigma-field induced by the $e$-variables observed by time $t$, with observation indices $R_{i,j}$ and continuation indicators $\mathbf{1}[j \leq N]$ observed by time $t+1$. Our filtration encodes the information before $e$-values at time $t+1$ are generated, but observability and continuation at time $t+1$ are known. By definition of $\Fcal_t$, the trajectory length $N$ is a potentially random stopping time adapted to $(\Fcal_t)_{t \in \NN}$, and the number of decisions, $M = \max\{i : \exists t \leq N \text{ such that } R_{i,t} = 1\}$,  is also allowed to be random. We let $\Fcal_N = \{A \in \Fcal_\infty: A \cap \{N=t\} \in \Fcal_t \ \forall t \geq 0\}$ denote the sigma-field corresponding to the final observed outputs, where $\Fcal_\infty = \sigma\left(\cup_{t=1}^\infty \Fcal_t\right)$. 
Write
\(
F_N=\bigl(M,N,(\mathbf e_i,\mathbf R_i)_{i=1}^{M}\bigr)
\)
for the observed trajectory and let \(\mathcal F_N=\sigma(F_N)\). Our data-driven weights are formally defined as \(c_i=c_i(F_N)\), with \(c_i=0\) whenever \(i>M\). 


Let \(\mathcal Q\) be the class of all joint distributions \(Q\) of the trajectory and correctness label variables
\(
(F_N,(\theta_i)_{i=1}^{M})
\)
that satisfy the following four assumptions:

\begin{assumption}[Finite Expected Number of Changes]\label{assump:finite_expected_changes}
    The number of changes/decisions $M$ satisfies $\EE_{Q}[M] < \infty$ and $M \geq 1$ $Q$-almost surely. 
\end{assumption}

\begin{assumption}[Finite Trajectory Length]\label{assump:finite_trajectory_length}
    $N$ is finite $Q$-almost-surely. 
\end{assumption}

\begin{assumption}[$e$-value Constraints]\label{assump:e_value_constraint}
    The pair $(e_{i,t}, \bar{e}_{i,t})$ are nonnegative $Q$-almost-surely. Furthermore, the expectation of the cumulative sum over $(1+e_{i,t}+\bar{e}_{i,t})$ is finite, i.e.,
    \begin{equation}\textstyle
        \EE_Q\left[ \sum_{i=1}^M\sum_{t=1}^N R_{i,t}(1+e_{i,t} + \bar{e}_{i,t}) \right] < \infty, 
    \end{equation}
    and their conditional expectation satisfies the $\Fcal_{t-1}$-conditional $e$-value constraint
    \begin{equation}
        \EE_{Q}\left[\theta _iR_{i,t}(e_{i,t} - 1) \mid \Fcal_{t-1}\right] \leq 0, \qquad \EE_Q\left[(1-\theta_i) R_{i,t}  (\bar{e}_{i,t} - 1) \mid \Fcal_{t-1} \right] \leq 0.
    \end{equation}
\end{assumption}
\begin{assumption}[Nonnegative Weights with Finite Expectation]\label{assump:target_param_restrictions}
     Target weights $c_i$ satisfy $c_i\geq0$ for $i\in[M]$ $Q$-almost-surely, and
     $\EE_Q\left[\sum_{i=1}^M c_i\right]<\infty$. 
\end{assumption}

For $Q \in \Qcal$, we let $\tilde{Q}$ denote its marginalized distribution of just the observed trajectory $F_N$, and let $\tilde\Qcal=\{\tilde Q:Q\in\Qcal\}$ denote the set of all such marginalized distributions.

The second condition of Assumption \ref{assump:e_value_constraint} states that $(e_{i,t}, \bar{e}_{i,t})$ are $\Fcal_{t-1}$-conditional $e$-values for the hypotheses $\theta_i = 1$ and $\theta_i = 0$ respectively \citep{vovk2021evalues}. The expectation contains $\theta_i$, as $\theta_i$ is a random label generated by the agent's actions.

\begin{example}[$e$-values for a binary goal metric]
Consider Example~\ref{ex:code_agent} with $X_{\mathrm{goal}}\in\{0,1\}$ and
\(
\theta_i=\mathbf{1}\!\left\{
\mathbb{E}_{P_i}[X_{\mathrm{goal}}]
\ge \mathbb{E}_{P_{j_i}}[X_{\mathrm{goal}}]
\right\}.
\)
Fix $0<b<a<1$ and let $r=(a+b)/2$.
At each evaluation time $t$ for decision $i$, observe one new draw $X_{\mathrm{goal},t}$ from $P_i$ and one $X'_{\mathrm{goal},t}$ from $P_{j_i}$, independently
conditional on the past. Let our $e$-values be the likelihood ratios
\[
e_{i,t}=\frac{\ell_t(b,a)}{\ell_t(r,r)},
\quad
\bar e_{i,t}=\frac{\ell_t(a,b)}{\ell_t(r,r)},\quad\text{where}\quad\ell_t(u,v)
=u^{X_{\mathrm{goal},t}}(1-u)^{1-X_{\mathrm{goal},t}}v^{X'_{\mathrm{goal},t}}(1-v)^{1-X'_{\mathrm{goal},t}}.
\]
The first has conditional expectation at most one when $\theta_i=1$,
and the second when $\theta_i=0$. Thus, large $e_{i,t}$ provides
evidence against correctness, while large $\bar e_{i,t}$ provides
evidence in its favor. Products of these likelihood ratios
accumulate evidence over successive evaluations. 
\end{example}

\subsection{Estimation and Inference on Agent Quality}

We construct classes of upper and lower bound estimators that bound $\theta_*$ in expectation uniformly over all distributions $Q \in \Qcal$. Our choice to construct estimators that bound $\theta_*$ in expectation is motivated by the fact that two distributions $Q$ and $Q'$ with equivalent marginalized distribution $\tilde{Q}$ can yield different expected values of $\theta_*$.
%
%
Specifically, our work aims to characterize the class of upper and lower bound estimators that provide uniform guarantees over all data generating processes $Q \in \Qcal$.

\begin{definition}[Uniformly Valid Bound Estimator]\label{defn:admissible_est}
    Let $\widehat\Theta_N$ denote the set of $\Fcal_N$-measurable, real-valued functions with finite expectation for all $Q \in \Qcal$. The function $\hat\theta_U \in \widehat\Theta_N$ is uniformly valid upper bound estimator if $\inf_{Q \in \Qcal}\EE_{Q}\left[ \hat\theta_U - \theta_*\right]  \geq 0$. 
\end{definition}
Uniformly valid lower bound
estimators are defined analogously, with the inequalities reversed. We also construct uniform confidence sets over $Q \in \Qcal$. 
\begin{definition}[Uniformly Valid Confidence Set]\label{defn:admissible_inf}
    Let $\Ccal_N$ denote the set of mappings from a fixed coverage error parameter $\alpha \in (0,1)$ and $F_N$ to a set on the real line. The mapping $C(\alpha, F_N)$ is a uniform $(1-\alpha)$ confidence set for $\theta_*$ if $\sup_{Q \in \Qcal} P_{Q}\left( \theta_* \not\in C(\alpha, F_N) \right) \leq \alpha$. 
\end{definition}
Uniform validity states that our estimators and confidence intervals provide valid guarantees in expectation and in probability across all $Q \in \Qcal$, enabling statistically rigorous evaluation of agentic quality for any sequence of agent decisions.

\section{Estimation Bounds on Agent Quality}\label{sec:estimator}

We begin with our class of estimators for upper bounds on $\theta_*$, which combines the observed $e$-value pairs with predictable weights to construct upper bound estimates. We discuss analogous results for our lower bound estimator class in Appendix \ref{app:lower_bound}.

To construct our upper bound estimator, we start with hypothesis-specific terms that have expectation upper bounding the correctness parameter $\theta_i$. For $i \in [M]$,
\begin{equation}\label{eq:e_processes}
    A_{i} = \sum_{t=1}^N w_{i,t} R_{i,t} (e_{i,t} - 1), \qquad \bar{A}_{i} = \sum_{t=1}^N \bar{w}_{i,t} R_{i,t}(\bar{e}_{i,t}- 1),
\end{equation}
where $w_{i,t}$ and $\bar{w}_{i,t}$ denote nonnegative $\Fcal_{t-1}$-measurable weights. Conditional on $\theta_i = 1$, the term $A_{i}$ has non-positive expectation along the evaluation trajectory. Thus, positive expected values of $A_{i}$ can only occur when $\theta_i=0$, and large values of $A_i$ yield evidence that $\theta_i = 0$. Likewise, larger values of $\bar{A}_{i}$ yields evidence that $\theta_i = 1$. We combine evidence on the correctness label $\theta_i$ with $\Fcal_N$-measurable weights $c_i$ using $\psi_i$, where 
\begin{equation}
    \psi_{i} = \max(c_i-A_{i}, -\bar{A}_{i}).
\end{equation}
The term $\psi_{i,t}$ corresponds to an upwards biased estimate of $c_i\theta_i$. Conditional on $\theta_i = 1$, the term $c_i-A_i$ has expectation at least $\EE_{\tilde{Q}}[c_i]$; conditional on $\theta_i = 0$, the term $(-\bar{A}_i)$ has nonnegative expectation. Therefore, $\psi_i$ has expectation at least $\EE_{\tilde{Q}}[c_i\theta_i]$. Our estimator is defined as the sum of $\psi_i$ over the $M$ agent decisions, 
\begin{equation}\label{eq:estimator}
    \hat\theta_*^U = \sum_{i=1}^M  \psi_i.
\end{equation}
We define our estimator class as all estimators of the form in Equation \eqref{eq:estimator}, with predictable weight sequences satisfying the following conditions. 
\begin{assumption}[Integrable Predictable Weights]\label{assump:integrable_pred_weights}
    For any $Q \in {\Qcal}$, weights $(w_{i,t}, \bar{w}_{i,t})$ are nonnegative, $\Fcal_{t-1}$-measurable, and equal zero whenever $R_{i,t} = 0$. Furthermore, the expectation of the sum of weights $\EE_{Q}\left[ \sum_{i=1}^M \sum_{t=1}^N R_{i,t}(w_{i,t} + \bar{w}_{i,t}) \right]$ is finite for all $Q \in \Qcal$. 
\end{assumption}
In particular, Assumption \ref{assump:integrable_pred_weights} imposes integrability conditions on our weights that ensure $\EE_{Q}\left[\hat\theta_*^U\right]$ is well-defined. For any choice of weights satisfying Assumption \ref{assump:integrable_pred_weights}, Theorem \ref{thm:admiss_est} shows that our estimator satisfies uniform validity over all $Q \in \Qcal$.

\begin{theorem}[Uniformly Valid Estimator]\label{thm:admiss_est}
    Let Assumptions \ref{assump:finite_expected_changes}--\ref{assump:target_param_restrictions} hold. Under any sequence of weights $(w_{i,t}, \bar{w}_{i,t})$ satisfying Assumption \ref{assump:integrable_pred_weights}, $\hat\theta_*^U$ satisfies Definition \ref{defn:admissible_est}. 
\end{theorem}


\subsection{Instance-wise Sharpness}

Uniform validity does not ensure that a procedure is informative: for example, $\sum_{i=1}^{M}c_i$ is a uniformly valid upper bound estimators. A natural notion of performance is the \textit{sharpness} of its expectation relative to $\tilde{Q}$, the distribution of observed outputs. Let $\Pi(\tilde{Q})\subseteq \Qcal$ correspond to the set of all agent trajectory distributions $Q$ with the marginalized distribution $\tilde{Q}$ over observed outputs $F_N$. An estimator $\hat\theta^U$ is a $\tilde{Q}$-sharp upper bound if its expectation equals largest possible expected value of $\theta_*$ over $\Pi(\tilde{Q})$,
\begin{equation}\label{eq:sharp_bound}
    \EE_{\tilde{Q}}\left[ \hat\theta^U \right] = U(\tilde{Q}), \qquad U(\tilde{Q}) = \sup_{Q \in \Pi(\tilde{Q})} \EE_{Q}[\theta_*]. 
\end{equation}
Below, we show that there exists a sequence of predictable weights such that our estimator $\hat\theta_*^U$ achieves arbitrarily close expectation equivalent to the $\tilde{Q}$-sharp upper bounds. 

\begin{theorem}[Sharpness over Observable Distributions]\label{thm:oracle_sharpness}
Let $U(\tilde{Q})$ be as in Equation \eqref{eq:sharp_bound}. Then, for any $\tilde{Q} \in \tilde{\Qcal}$ and $\epsilon >0$, there exists a sequence of weights $(w_{i,t}, \bar{w}_{i,t})$ satisfying Assumption \ref{assump:integrable_pred_weights} such that $\hat\theta_*^U$ satisfies $\EE_{\tilde{Q}}\left[\hat\theta_*^U\right] \leq  U(\tilde{Q}) + \epsilon$.  
\end{theorem}


Theorem \ref{thm:oracle_sharpness} shows that our class is arbitrarily close to the sharp bound $U(\tilde{Q})$ for any $\tilde{Q} \in \tilde{\Qcal}$. Paired with Theorem \ref{thm:admiss_est}, our theoretical results demonstrate that uniform validity does not come at the cost of instance-wise sharpness: the arbitrarily sharp estimator defined by Theorem \ref{thm:oracle_sharpness} lives in our estimator class, and therefore satisfies uniform validity. 

\subsection{Comparisons with Storey's Estimator}

To compare our approach with existing methods for estimating valid upper bounds, we consider an adaptation of the classical Storey estimator \citep{storey2002direct}, which is based on $p$-values rather than $e$-values. Storey's estimator $\pi_m^{\text{Storey}}(\lambda)$ is given by 
\begin{equation}\label{eq:storey}
    \pi^{\text{Storey}}(\lambda) = \left(\frac{\sum_{i=1}^M \mathbf{1}\left[ p_i \geq  \lambda\right] }{M(1-\lambda)}\right),
\end{equation}
and corresponds to expected upper bound for the proportion of correct hypotheses under super-uniform $p$-values. The parameter $\lambda \in [0,1)$ is specified in advance. To compare our proposed estimators with existing results, we focus on the setting where $M$ is fixed, and we observe a single $e$-value $e_i$ for each $i \in [M]$. Consider the goal of estimating $\theta_* = \frac{1}{M}\sum_{i=1}^M \theta_i$. In our fixed $M$ setting, an upper bound estimator can be obtained by converting our $e$-values into $p$-values and leveraging Storey's estimator. First, by \cite{ramdas2023gametheoreticstatisticssafeanytimevalid}, we convert $e$-values $(e_1, \dots, e_M)$ into $p$-values with the dominating $e$-to-$p$ converter $p_i = \min(1, 1/e_i)$. Then, Storey's estimator yields  
\begin{equation}
     \pi^{\text{Storey}}(\lambda) = \left(\frac{\sum_{i=1}^M \mathbf{1}\left[ e_i \leq  1/\lambda\right] }{M(1-\lambda)}\right).
\end{equation}
Storey's estimator with $e$-to-$p$ conversion obtains expectation larger than $\theta_*$ in our simplified setting, yielding a valid upper bound estimator. We show that for any choice of $\lambda \in (0,1)$, there exists a choice of weight for our estimator $\hat\theta_*^U$ which yields estimated upper bounds no larger than Storey's estimator in a a pathwise sense. 

\begin{lemma}[Dominating Choice for Correctness Proportion]\label{lem:domnating}
    Let $\lambda \in (0,1)$ be fixed, with $M$ fixed and a single $e$-value $(e_i)_{i=1}^M$ for each $i \in [M]$. Let $\bar{w}_i = 0$ for all $i \in M$, yielding 
    \begin{equation}
        \hat\theta_*^U = \sum_{i=1}^M \max\left(\frac{1}{M} - w_i(e_i - 1), 0\right). 
    \end{equation}
    Then, $w_i =  \frac{\lambda}{M(1-\lambda)}$ yields that $\hat\theta_*^U \leq \pi^{\text{Storey}}(\lambda)$ for any realization of $e$-values $(e_1, \dots, e_M)$.  
\end{lemma}

Thus, in the $e$-value setting, a standard adaptation of Storey's approach always dominated by a member of our estimator class with simple, fixed weights.

\begin{remark}[Leveraging Lemma \ref{lem:domnating} in Practice.]
    To leverage the results of Lemma \ref{lem:domnating} under more general settings, one can merge sequences of $(e_{i,t}, \bar{e}_{i,t})_{t \in \NN}$ according to the martingale merging function of \cite{vovk2024mergingsequentialevaluesmartingales} in Equation \eqref{eq:vovk_wang} to obtain a single pair of $e$-values $(e_i, \bar{e}_i)$. For example, with the choice of $\lambda_{i,t} = 1$, $(e_i, \bar{e}_i)$ corresponds to
    \begin{equation}
        e_i = \prod_{t \in [N]: R_{i,t} = 1} e_{i,t}, \qquad \bar{e}_i = \prod_{t \in [N]: R_{i,t} = 1} \bar{e}_{i,t},
    \end{equation}
    the product $e$-variable. By setting $w_i = \lambda/(M(1-\lambda))$ and $\bar{w}_i = 0$, our final estimator is no worse than conducting $e$-to-$p$ conversion and applying Storey's estimator. 
\end{remark}

\section{Confidence Sets for Agent Quality}\label{sec:inference}

Our confidence sets build upon our estimation framework by leveraging terms $(A_i, \bar{A}_i)$ as measures of evidence. Let $\mathcal A_t=\{i: R_{i,t}=1\}$ as the changes with evidence observed at time $t$, and
denote running versions of the evidence terms as $A_{i,t}$ and $\bar{A}_{i,t}$, defined as 
\begin{equation}\label{eq:running}
  A_{i,t}=\sum_{j=1}^{t} w_{i,j}R_{i,j}(e_{i,j}-1),\qquad
  \bar A_{i,t}=\sum_{j=1}^{t} \bar w_{i,j}R_{i,j}(\bar e_{i,j}-1),
\end{equation}
with $A_{i,0} = \bar{A}_{i,0} = 0$. For candidate label vectors $v\in\{0,1\}^M$, we
define \emph{candidate wealth}
\begin{equation}\label{eq:cand-wealth}
  K_t(\vartheta)=1+\sum_{i=1}^M\big(\vartheta_iA_{i,t}+(1-\vartheta_i)\bar A_{i,t}\big).
\end{equation}
When $\vartheta=(\theta_1,\dots,\theta_M)$, each increment of $K_t(v)$ has non-positive $\mathcal F_{t-1}$-conditional
expectation by Assumption \ref{assump:e_value_constraint}, so $K_t(\theta)$ starts at wealth $1$ and drifts downward. Thus, large values of
$K_t(v)$ are evidence against the vector $v = (\theta_1, \dots, \theta_M)$. Our confidence set $C(\alpha, F_N)$ is the projection of all label vectors $v \in \{0,1\}^{M}$ with wealth less than $1/\alpha$, i.e. 
\begin{equation}\label{eq:confidence_interval}
    C(\alpha, F_N) = \left\{ \sum_{i=1}^M c_iv_i : v \in V(\alpha, F_N) \right\}, \quad V(\alpha, F_N)= \left\{ v \in \{0,1\}^{M}: K_N(v) < \frac{1}{\alpha} \right\}.
\end{equation}
If $K_N(v)$ is nonnegative, it forms a valid $e$-value for $(\theta_1, \dots, \theta_M) = \vartheta$ by Assumption \ref{assump:e_value_constraint}. Our confidence set simply corresponds to the collection of all vectors $v$ that fail to be rejected. We provide restrictions on predictable weight sequences that ensure $K_N(v)$ is a valid $e$-value for all $v \in \{0,1\}^M$ uniformly over $Q \in \Qcal$. The largest and smallest elements of
$C(\alpha, F_N)$ maximize and minimize the linear objective $\sum_{i} c_i v_i$ over binary $v$
subject to the single linear constraint $K_N(v) < 1/\alpha$, which is a 0--1 knapsack problem. When the coefficients are equal, $c_i = c$, no solver is needed.

\subsection{Sequential Betting with Shared Wealth}


To characterize our predictable weight restrictions, we first parameterize $K_t(\vartheta)$ as
\begin{align}
    &K_t(\vartheta) = B_t + \sum_{i=1}^M (\vartheta_i D_{i,t} + (1-\vartheta_i)\bar{D}_{i,t}), \label{eq:16}\\
    B_t = 1 + \sum_{i=1}^M &\min\{A_{i, t}, \bar{A}_{i, t} \}, \quad D_{i,t} = (A_{i,t} - \bar{A}_{i,t})_+, \quad \bar{D}_{i,t} = (\bar{A}_{i,t} - A_{i,t})_+. 
\end{align}
The value $B_t$ denotes the wealth common to every candidate vector $\vartheta \in \{0,1\}^M$, while $D_{i,t}$ and $\bar{D}_{i,t}$ denote the excess evidence against $\theta_i = 1$ and $\theta_i = 0$ respectively. Then, $K_t(\vartheta)$ corresponds to the common wealth pool $B_t$ plus the excess evidence  on the side that $\bar\theta$ bets against. Our class of inference procedures restricts weights $(w_{i,t}, \bar{w}_{i,t})$ as follows. 

\begin{assumption}[Sequential Bets with Shared Wealth]\label{assump:betting_and_transfers}
For every $t\in\mathbb N$, $Q \in \Qcal$, weights $w_{i,t}$ and $\bar{w}_{i,t}$ can be represented as $w_{i,t}=\lambda_{i,t}\big(D_{i,t-1}+b_{i,t}\big)$ and $\bar{w}_{i,t}=\bar\lambda_{i,t}\big(\bar D_{i,t-1}+b_{i,t}\big)$,
where (i) the betting fractions $\lambda_{i,t},\bar\lambda_{i,t}\in[0,1]$ and the transfers
$b_{i,t}\ge0$ are $\mathcal F_{t-1}$-measurable, (ii) equal $b_{i,t}= 0$ and $\lambda_{i,t} = \bar{\lambda}_{i,t} = 0$ whenever $R_{i,t}=0$, and (iii) the transfers $b_{i,t}$
satisfy the budget constraint $\sum_{i\in\mathcal A_t} b_{i,t}\leq B_{t-1}$ $Q$-almost surely.
\end{assumption}
Assumption \ref{assump:betting_and_transfers} parametrizes weights $(w_{i,t}, \bar{w}_{i,t})$ with betting fractions $(\lambda_{i,t}, \bar{\lambda}_{i,t})$ and wealth transfer parameters $b_{i,t}$. By the budget constraint $\sum_{i \in \Acal_t} B_{t-1}$, note that $K_t(v)$ is nonnegative for all $t\in \NN$. The transfer parameter $b_{i,t}$ denotes the amount of wealth $B_{t-1}$ allocated to testing $\theta_i$'s value. Larger values of $b_{i,t}$ correspond to larger possible wealth allocation, reflected in the upper bounds of $(w_{i,t}, \bar{w}_{i,t})$. Thus, larger transfers to testing $\theta_i$ correspond to larger weight upper bounds in our construction. The fractions $(\lambda_{i,t}, \bar{\lambda}_{i,t})$ are the usual betting fractions of a game martingale \citep{vovk2024mergingsequentialevaluesmartingales}, determining how much capital is invested for collecting evidence against $\theta_i = 1$ and $\theta_i = 0$ relative to its hypothesis specific budget $D_{i, t-1} + b_{i,t}$ and $\bar{D}_{i, t-1} + b_{i,t}$ respectively.



\paragraph{Recovering $e$-Value Merging}
Let $(E_{i,N},\bar{E}_{i,N})$ denote the $e$-values for $\theta_i = 1$ and $\theta_i = 0$, merged with the game-martingale of \cite{vovk2024mergingsequentialevaluesmartingales}, 
\begin{equation}\label{eq:vovk_wang}
    E_{i,N} = \prod_{j \leq t: R_{i,j} = 1} (1+ \lambda_{i,j} (e_{i,j}-1)), \qquad \bar{E}_{i,t} =  \prod_{j \leq t: R_{i,j} = 1} (1+ \bar\lambda_{i,j} (\bar{e}_{i,j}-1)).
\end{equation}
By Assumption \ref{assump:e_value_constraint}, martingale merging yields valid $e$-values $E_{i,N}$ and $\bar{E}_{i,N}$ for $\theta_i = 1$ and $\theta_i = 0$. For a fixed, nonnegative vector $b \in \RR^{M+1}$ with $\sum_{i=0}^M b_i = 1$, the weighted average merging for arbitrarily dependent hypotheses \citep{wang2025admissiblewaymergingarbitrary} yields an $e$-value for $(\theta_1, \dots, \theta_M) = v$, given by $K_N(\vartheta) = \left(b_0+\sum_{i=1}^Mb_i\left(\vartheta_i E_{i,N} + (1-\vartheta_i) \bar{E}_{i,N}   \right)\right)$. Setting $(\lambda_{i,j}, \bar{\lambda}_{i,j})$ as above, $ b_{i,1} = b_i$ at $t=1$ and $b_{i,t} = b_{i,1}\min(E_{i, t-1}, \bar{E}_{i, t-1})$ recovers this construction. Note that Assumption \ref{assump:betting_and_transfers} enables a more general construction: $(b_{i,t})$ are allowed to pick any allocation among a common wealth pool $B_{t-1}$ satisfying budget constraints.  


\paragraph{Restrictions under Random Hypothesis Labels}
Assumption \ref{assump:betting_and_transfers} requires an additional restriction relative to typical confidence sets: every $\vartheta\in\{0,1\}^M$ with $\vartheta_i=1$ contains the
same term $A_{i,t}$ in $K_t(\vartheta)$, and every $\vartheta$ with $\vartheta_i=0$ contains the same term $\bar A_{i,t}$ due to \textit{shared} weights $(w_{i,t},\bar w_{i,t})$ and wealth pool $B_t$ across all $v \in \{0,1\}^M$. In contrast, when $\theta_i$ are fixed, each label vector $\vartheta$ can have its own, hypothesis specific weights $(w_{i,t}^\vartheta, \bar{w}_{i,t}^\vartheta)$, wealth $B_t^\vartheta$, and transfers $b_{i,t}^\vartheta$. Failure of hypothesis-specific tuning parameters can be shown by selecting $Q$ with dependence between $e$-values and labels $\theta_i$.



\begin{example}[An agent reacting to a false alarm]\label{ex:shared-stakes}
Let $\alpha=1/10$, $M=2$, $N=1$, $c_1=c_2=1$ and $\bar e_1=\bar e_2=1$. First, an agent ships a correct change $\theta_1=1$, with $e_1=10$ with probability $1/10$, a false alarm, and $e_1=0$
otherwise. After a false alarm $(e_1 = 10)$, it ships a fix that is bad
($\theta_2=0$); otherwise it ships a good change ($\theta_2=1$). Change 2's own experiment yields $e_2=20$ when $\theta_2 = 0$, and $e_2=20$ with probability $1/20$ and $e_2=0$ otherwise. Both $e$-values satisfy Assumption \ref{assump:e_value_constraint}. Consider separate weights for each $v$:
\begin{equation}
    K_N(1,1) = (e_1 + e_2)/2, \quad K_N(1,0)= e_1, \quad K_N(0,1) = e_2, \quad K_N(0,0) = 1.
\end{equation}
Under this trajectory distribution, separate weights fail to cover $\theta_*$ with probability $0.145 > \alpha$. Shared weights $w_{i,1} = \bar{w}_{i,1} = 1/2$ for $i\in[M]$ fail to cover $\theta_*$ with probability $0.045 < \alpha$. 

\end{example}

\subsection{Uniform Coverage over Agent Trajectories}

Under our regularity conditions in Section \ref{sec:setup} and the sequential betting restrictions of Assumption \ref{assump:betting_and_transfers}, we obtain our uniform coverage guarantee in Theorem \ref{thm:coverage}. 

\begin{theorem}[Uniform Confidence Sets]\label{thm:coverage}
    Let $\alpha \in (0,1)$ be fixed, and assume Assumptions \ref{assump:finite_expected_changes}--\ref{assump:target_param_restrictions} hold. Under any weight sequences $(w_{i,t}, \bar{w}_{i,t})$ satisfying Assumption \ref{assump:betting_and_transfers}, the confidence set $C(\alpha, F_N)$ in Equation \eqref{eq:confidence_interval} is a uniformly valid $(1-\alpha)$ confidence set (Definition \ref{defn:admissible_inf}).
\end{theorem}

\section{Case Study: Comparing Coding Agents}\label{sec:experiments}
For our real-world case study, we compare the performance of two large language models for generating code speed-ups on the \texttt{NetworkX} Python library \citep{SciPyProceedings_11}.

\paragraph{Setup} We evaluate Claude Opus 5 and Claude Haiku 4.5 run at high effort and asked to propose up to 50 localized runtime optimizations to the \texttt{NetworkX} package. Each agent was run separately. Changes were proposed sequentially on the code tree left by the agent’s previously accepted changes. Along each agent's trajectory, change $i$ was correct if the geometric-mean runtime decreases by at least $p\%$ from the last change. We set $c_i = 1/M$.

\paragraph{Random Stopping and $e$-Values}

For each agent, its $i$-th change records the contrasts $(Y_{i,1},\ldots,Y_{i,N_i})$ over $N_i$ generated graphs, where $Y_{i,t}$ is the log runtime for change $i$ minus the log runtime for change $i-1$. The graphs used for evaluations use seeds unavailable to either agent, corresponding to an unknown distribution of user inputs. For a minimum speedup of $p\%$, we set $\delta_p=-\log(1-p/100)$ and apply the one-sided right-Haar construction of Appendix \ref{app:eprocess} to the shifted contrasts $X_{i,t}=Y_{i,t}+\delta_p$. This produces $E_{i,t}$ and $\bar E_{i,t}$, with increment pairs $e_{i,t}=E_{i,t}/E_{i,t-1}$ and $\bar e_{i,t}=\bar E_{i,t}/\bar E_{i,t-1}$, providing evidence against $\theta_i=1$ and $\theta_i=0$, respectively. The sample size $N_i$ is determined adaptively. After the first 10 pairs, stopping is assessed every five additional pairs; sampling stops when either $E_{i,t}$ or $\bar{E}_{i,t}$ reaches 100 or when 150 pairs have been observed.

\paragraph{Predictable Tuning.} Across all experiments, we use a predictable tuning heuristic based on trajectory simulation at each time $t$ to select weights $(w_{i,t}, \bar{w}_{i,t})$. For our estimator, our approach aims to minimize and maximize the conditional expectation of the upper and lower bound estimators. For inference, we similarly minimize a convex upper bound on the expected width. Both methods leverage 64 trajectory simulations at each time $t$ to approximate conditional expectations at each $t \in \NN$ to obtain weights $(w_{i,t}, \bar{w}_{i,t})$. Our approach, along with its pseudocode, is provided in Appendix \ref{app:tuning}.

\subsection{Discussion of Results}

\begin{figure}[t]
    \centering
    \includegraphics[width=\linewidth]{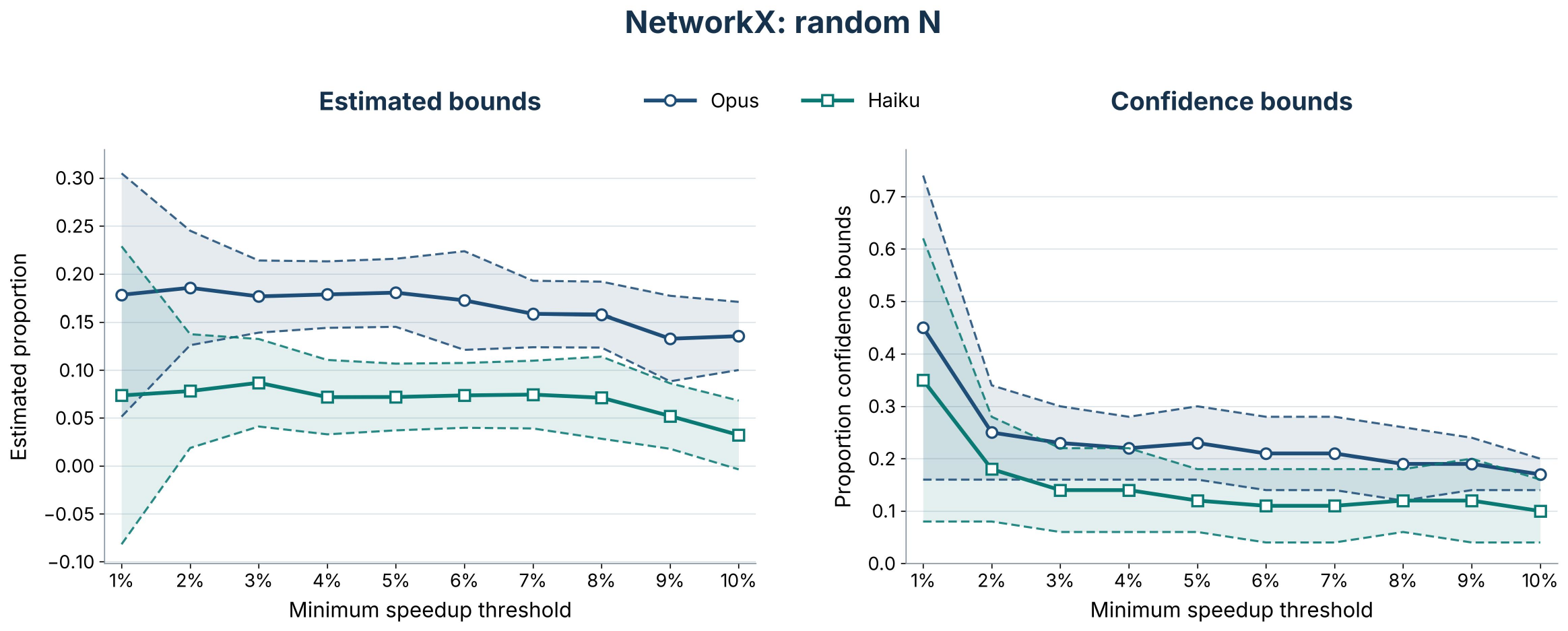}
    \caption{Estimated Bounds and Confidence Intervals for \texttt{NetworkX} Experiment. Confidence sets are pointwise for each minimum speedup threshold and agent, with $\alpha = 0.1$. The bolded line denotes the midpoint of our estimated bounds and confidence sets.}
    \label{fig:the_only_one}
\end{figure}

In Figure \ref{fig:the_only_one}, we plot our estimated upper/lower bound estimators and confidence sets as a function of $p$, the minimum speedup threshold required for a change to be considered correct. Across all thresholds, Opus exhibits a higher estimated proportion of correct changes than Haiku. For example, at a minimum speedup of 5\%, the lower estimated bound is approximately 0.15 for Opus, while the estimated upper bound for Haiku is 0.11. The gap persists as the minimum speedup threshold increases, suggesting that Opus produces more meaningful runtime improvements than Haiku.  

Our confidence sets provide additional information about the reliability of these conclusions. Despite adaptively determined sample sizes and dependence between successive code changes, our procedures yield informative 90\% pointwise confidence sets across the considered speedup thresholds. In particular, the lower confidence sets remain positive even when requiring a 10\% speedup, providing statistical evidence that both agents produce a nonzero proportion of meaningful improvements.

Our case study demonstrates that that agent quality can be meaningfully evaluated on tasks without ground-truth correctness labels, while preserving rigorous statistical guarantees. In our experiment, the true geometric-mean runtime improvement of each code change is unknown, and evidence is collected and stopped adaptively. Our framework converts this sequential evidence into interpretable estimates and confidence bounds on a user-defined notion of agent quality, such as the fraction of beneficial decisions. For practitioners choosing between agents, metrics such as the proportion of correct changes enable comparisons based on the frequency of practically meaningful improvements, rather than benchmark performance or noisy empirical correctness labels. Additionally, correctness, such as minimum improvement thresholds, can be selected to reflect deployment priorities, allowing practitioners to define correctness based on operational significance. 

\section{Conclusions}

As AI agents increasingly make consequential decisions in real-world systems, evaluating their performance requires moving beyond static benchmarks or heuristic correctness labels. We introduced a null-proportion estimation framework for evaluating agent quality based on sequential $e$-values, allowing for adaptive decisions, data-dependent stopping, and quality metrics defined after deployment. Our results establish that uniformly valid estimation and inference are possible under these conditions, with estimators that can approach instance-wise sharp bounds. Our coding agent case study illustrates how this framework can provide interpretable assessments of agent performance.

Our work opens several directions for future research. Our setting is intentionally general, and we construct uniform estimated bounds and confidence sets. While our approach ensures our desired statistical guarantees across any distribution of agent trajectories, more efficient procedures for estimation and inference may be possible under stronger assumptions, such i.i.d. $e$-value pairs, a fixed data collection policy, or restrictions on agent behavior. A practical direction of interest is the development of more efficient tuning procedures, including improved predictable methods for allocating evidence across hypotheses and over time. Lastly, our experiment is intended as a case study, rather than a rigorous evaluation across agents, tasks, and correctness objectives. Because our methods have uniform guarantees, testing them on large scale evaluation across recorded agent trajectories offers the promise of selecting agents in a statistically principled manner based past performance, tailored to the relevant task.


\section*{Data Availability Statement}

The data and code needed to reproduce the numerical results are provided with this submission. The starting version of \texttt{NetworkX} for the case study is available at \url{https://github.com/networkx/networkx} (commit~\texttt{2622da75c}).

\section*{Acknowledgements}

Generative AI tools were used to edit this manuscript, check mathematical arguments, and write code for experiments. The authors reviewed all such output and take full responsibility for the content.

\section*{Conflicts of interest}

The authors declare no conflicts of interest.

\bibliographystyle{abbrvnat}
\bibliography{main}

@misc{alshahwan2024automatedunittestimprovement,
      title={Automated Unit Test Improvement using Large Language Models at Meta}, 
      author={Nadia Alshahwan and Jubin Chheda and Anastasia Finegenova and Beliz Gokkaya and Mark Harman and Inna Harper and Alexandru Marginean and Shubho Sengupta and Eddy Wang},
      year={2024},
      eprint={2402.09171},
      archivePrefix={arXiv},
      primaryClass={cs.SE},
      url={https://arxiv.org/abs/2402.09171}, 
}

@misc{novikov2025alphaevolvecodingagentscientific,
      title={AlphaEvolve: A coding agent for scientific and algorithmic discovery}, 
      author={Alexander Novikov and Ngân Vũ and Marvin Eisenberger and Emilien Dupont and Po-Sen Huang and Adam Zsolt Wagner and Sergey Shirobokov and Borislav Kozlovskii and Francisco J. R. Ruiz and Abbas Mehrabian and M. Pawan Kumar and Abigail See and Swarat Chaudhuri and George Holland and Alex Davies and Sebastian Nowozin and Pushmeet Kohli and Matej Balog},
      year={2025},
      eprint={2506.13131},
      archivePrefix={arXiv},
      primaryClass={cs.AI},
      url={https://arxiv.org/abs/2506.13131}, 
}

@article{Gottweis_2026,
   title={Accelerating scientific discovery with Co-Scientist},
   volume={655},
   ISSN={1476-4687},
   url={http://dx.doi.org/10.1038/s41586-026-10644-y},
   DOI={10.1038/s41586-026-10644-y},
   number={8122},
   journal={Nature},
   publisher={Springer Science and Business Media LLC},
   author={Gottweis, Juraj and Weng, Wei-Hung and Daryin, Alexander and Tu, Tao and Sirkovic, Petar and Myaskovsky, Artiom and Glowaty, Grzegorz and Weissenberger, Felix and Orlandi, Alessio and Popovici, Dan and Palepu, Anil and Rong, Keran and Tanno, Ryutaro and Saab, Khaled and Zhang, Fan and Blum, Jacob and Carroll, Andrew and Kulkarni, Kavita and Tomašev, Nenad and Zverinski, Dina and Rendulic, Ivor and Vedadi, Elahe and Hasler, Florian and Rimanic, Luka and Boia, Marina and Budiselic, Ivan and Feinstein, Ben and Bellaiche, Mathias and Sheffer, Tom and Freyberg, Jan and Ratcliff, Jeremy and Bertolli, Ottavia and Chou, Katherine and Hassidim, Avinatan and Gokturk, Burak and Vahdat, Amin and Guan, Yuan and Dhillon, Vikram and Vaishnav, Eeshit Dhaval and Lee, Byron and Costa, Tiago R. D. and Penadés, José R. and Peltz, Gary and Matias, Yossi and Manyika, James and Hassabis, Demis and Xu, Yunhan and Kohli, Pushmeet and Pawlosky, Annalisa and Karthikesalingam, Alan and Natarajan, Vivek},
   year={2026},
   month=May, pages={487–496} }

@inproceedings{zheng2023judging,
  title     = {Judging {LLM}-as-a-Judge with {MT}-Bench and Chatbot Arena},
  author    = {Zheng, Lianmin and Chiang, Wei-Lin and Sheng, Ying and Zhuang, Siyuan and Wu, Zhanghao and Zhuang, Yonghao and Lin, Zi and Li, Zhuohan and Li, Dacheng and Xing, Eric P. and Zhang, Hao and Gonzalez, Joseph E. and Stoica, Ion},
  booktitle = {Advances in Neural Information Processing Systems (Datasets and Benchmarks Track)},
  volume    = {36},
  year      = {2023}
}

@article{wang2020methods,
  title   = {Methods for correcting inference based on outcomes predicted by machine learning},
  author  = {Wang, Siruo and McCormick, Tyler H. and Leek, Jeffrey T.},
  journal = {Proceedings of the National Academy of Sciences},
  volume  = {117},
  number  = {48},
  pages   = {30266--30275},
  year    = {2020},
  doi     = {10.1073/pnas.2001238117}
}

@article{angelopoulos2023ppi,
  title   = {Prediction-powered inference},
  author  = {Angelopoulos, Anastasios N. and Bates, Stephen and Fannjiang, Clara and Jordan, Michael I. and Zrnic, Tijana},
  journal = {Science},
  volume  = {382},
  number  = {6671},
  pages   = {669--674},
  year    = {2023}
}

@inproceedings{egami2023using,
  title     = {Using Imperfect Surrogates for Downstream Inference: Design-based Supervised Learning for Social Science Applications of Large Language Models},
  author    = {Egami, Naoki and Hinck, Musashi and Stewart, Brandon M. and Wei, Hanying},
  booktitle = {Advances in Neural Information Processing Systems},
  volume    = {36},
  year      = {2023}
}

@inproceedings{boyeau2025autoeval,
  title     = {{AutoEval} Done Right: Using Synthetic Data for Model Evaluation},
  author    = {Boyeau, Pierre and Angelopoulos, Anastasios Nikolas and Li, Tianle and Yosef, Nir and Malik, Jitendra and Jordan, Michael I.},
  booktitle = {Proceedings of the 42nd International Conference on Machine Learning},
  series    = {PMLR},
  volume    = {267},
  pages     = {5276--5290},
  year      = {2025}
}

@article{meinshausen2006estimating,
  title   = {Estimating the proportion of false null hypotheses among a large number of independently tested hypotheses},
  author  = {Meinshausen, Nicolai and Rice, John},
  journal = {The Annals of Statistics},
  volume  = {34},
  number  = {1},
  pages   = {373--393},
  year    = {2006}
}

@article{efron2008microarrays,
  title   = {Microarrays, Empirical {Bayes} and the Two-Groups Model},
  author  = {Efron, Bradley},
  journal = {Statistical Science},
  volume  = {23},
  number  = {1},
  pages   = {1--22},
  year    = {2008}
}

@article{ramdas2023gametheoretic,
  title   = {Game-theoretic statistics and safe anytime-valid inference},
  author  = {Ramdas, Aaditya and Gr{\"u}nwald, Peter and Vovk, Vladimir and Shafer, Glenn},
  journal = {Statistical Science},
  volume  = {38},
  number  = {4},
  pages   = {576--601},
  year    = {2023}
}

@article{johari2022always,
  title   = {Always Valid Inference: Continuous Monitoring of {A/B} Tests},
  author  = {Johari, Ramesh and Koomen, Pete and Pekelis, Leonid and Walsh, David},
  journal = {Operations Research},
  volume  = {70},
  number  = {3},
  pages   = {1806--1821},
  year    = {2022}
}

@inproceedings{lindon2022rapid,
  title     = {Rapid Regression Detection in Software Deployments through Sequential Testing},
  author    = {Lindon, Michael and Sanden, Chris and Shirikian, Vach{\'e}},
  booktitle = {Proceedings of the 28th ACM SIGKDD Conference on Knowledge Discovery and Data Mining},
  year      = {2022},
  doi       = {10.1145/3534678.3539099}
}

@article{sion1958general,
  author  = {Sion, Maurice},
  title   = {On general minimax theorems},
  journal = {Pacific Journal of Mathematics},
  volume  = {8},
  number  = {1},
  pages   = {171--176},
  year    = {1958},
  doi     = {10.2140/pjm.1958.8.171}
}

@article{alaoglu1940weak,
  author  = {Alaoglu, Leonidas},
  title   = {Weak Topologies of Normed Linear Spaces},
  journal = {Annals of Mathematics},
  volume  = {41},
  number  = {2},
  pages   = {252--267},
  year    = {1940},
  doi     = {10.2307/1968829}
}

@article{tychonoff1930topological,
  author  = {Tychonoff, Andrey},
  title   = {{\"U}ber die topologische Erweiterung von R{\"a}umen},
  journal = {Mathematische Annalen},
  volume  = {102},
  pages   = {544--561},
  year    = {1930},
  doi     = {10.1007/BF01782364}
}

@InProceedings{SciPyProceedings_11,
  author =       {Aric A. Hagberg and Daniel A. Schult and Pieter J. Swart},
  title =        {Exploring Network Structure, Dynamics, and Function using NetworkX},
  booktitle =   {Proceedings of the 7th Python in Science Conference},
  pages =     {11 - 15},
  address = {Pasadena, CA USA},
  year =      {2008},
  editor =    {Ga\"el Varoquaux and Travis Vaught and Jarrod Millman},
}

@misc{wang2025admissiblewaymergingarbitrary,
      title={The only admissible way of merging arbitrary e-values}, 
      author={Ruodu Wang},
      year={2025},
      eprint={2409.19888},
      archivePrefix={arXiv},
      primaryClass={math.ST},
      url={https://arxiv.org/abs/2409.19888}, 
}

@misc{jimenez2024swebenchlanguagemodelsresolve,
      title={SWE-bench: Can Language Models Resolve Real-World GitHub Issues?}, 
      author={Carlos E. Jimenez and John Yang and Alexander Wettig and Shunyu Yao and Kexin Pei and Ofir Press and Karthik Narasimhan},
      year={2024},
      eprint={2310.06770},
      archivePrefix={arXiv},
      primaryClass={cs.CL},
      url={https://arxiv.org/abs/2310.06770}, 
}

@misc{xie2026scientificobservationsmechanismsbenchmarking,
      title={From Scientific Observations to Mechanisms: Benchmarking Hypothesis Generation by AI Scientists}, 
      author={Xiaxun Xie and Qingqing Long and Meng Xiao and Wei Ju and Yuanchun Zhou and Xuezhi Wang and Hengshu Zhu},
      year={2026},
      eprint={2610.05197},
      archivePrefix={arXiv},
      primaryClass={cs.AI},
      url={https://arxiv.org/abs/2610.05197}, 
}

@misc{ramdas2023gametheoreticstatisticssafeanytimevalid,
      title={Game-theoretic statistics and safe anytime-valid inference}, 
      author={Aaditya Ramdas and Peter Grünwald and Vladimir Vovk and Glenn Shafer},
      year={2023},
      eprint={2210.01948},
      archivePrefix={arXiv},
      primaryClass={math.ST},
      url={https://arxiv.org/abs/2210.01948}, 
}

@misc{waudbysmith2024timeuniformcentrallimittheory,
      title={Time-uniform central limit theory and asymptotic confidence sequences}, 
      author={Ian Waudby-Smith and David Arbour and Ritwik Sinha and Edward H. Kennedy and Aaditya Ramdas},
      year={2024},
      eprint={2103.06476},
      archivePrefix={arXiv},
      primaryClass={math.ST},
      url={https://arxiv.org/abs/2103.06476}, 
}

@misc{bibaut2024nearoptimalnonparametricsequentialtests,
      title={Near-Optimal Non-Parametric Sequential Tests and Confidence Sequences with Possibly Dependent Observations}, 
      author={Aurelien Bibaut and Nathan Kallus and Michael Lindon},
      year={2024},
      eprint={2212.14411},
      archivePrefix={arXiv},
      primaryClass={stat.ME},
      url={https://arxiv.org/abs/2212.14411}, 
}

@article{storey2002direct,
  author    = {Storey, John D.},
  title     = {A direct approach to false discovery rates},
  journal   = {Journal of the Royal Statistical Society: Series B (Statistical Methodology)},
  volume    = {64},
  number    = {3},
  pages     = {479--498},
  year      = {2002},
  doi       = {10.1111/1467-9868.00346}
}

@article{vovk2021evalues,
  author    = {Vovk, Vladimir and Wang, Ruodu},
  title     = {E-values: Calibration, combination and applications},
  journal   = {The Annals of Statistics},
  volume    = {49},
  number    = {3},
  pages     = {1736--1754},
  year      = {2021},
  doi       = {10.1214/20-AOS2020}
}

@article{grunwald2024safe,
  author    = {Gr{\"u}nwald, Peter and de Heide, Rianne and Koolen, Wouter M.},
  title     = {Safe testing},
  journal   = {Journal of the Royal Statistical Society: Series B (Statistical Methodology)},
  volume    = {86},
  number    = {5},
  pages     = {1091--1128},
  year      = {2024},
  doi       = {10.1093/jrsssb/qkae011}
}

@misc{vovk2024mergingsequentialevaluesmartingales,
      title={Merging sequential e-values via martingales}, 
      author={Vladimir Vovk and Ruodu Wang},
      year={2024},
      eprint={2007.06382},
      archivePrefix={arXiv},
      primaryClass={math.ST},
      url={https://arxiv.org/abs/2007.06382}, 
}

@article{lai1976confidence,
  author  = {Lai, Tze Leung},
  title   = {On Confidence Sequences},
  journal = {The Annals of Statistics},
  volume  = {4},
  number  = {2},
  pages   = {265--280},
  year    = {1976}
}

@article{perezortiz2024estatistics,
  author  = {P{\'e}rez-Ortiz, Muriel Felipe and Lardy, Tyron and de Heide, Rianne and Gr{\"u}nwald, Peter D.},
  title   = {E-statistics, group invariance and anytime-valid testing},
  journal = {The Annals of Statistics},
  volume  = {52},
  number  = {4},
  pages   = {1410--1432},
  year    = {2024}
}

@article{wang2025anytime,
  author  = {Wang, Hongjian and Ramdas, Aaditya},
  title   = {Anytime-valid t-tests and confidence sequences for {G}aussian means with unknown variance},
  journal = {Sequential Analysis},
  year    = {2025},
  note    = {arXiv:2310.03722}
}

@article{lindon2025anytime,
  author  = {Lindon, Michael and Ham, Dae Woong and Tingley, Martin and Bojinov, Iavor},
  title   = {Anytime-Valid Linear Models and Regression Adjusted Causal Inference in Randomized Experiments},
  journal = {Journal of the American Statistical Association},
  year    = {2026},
  doi     = {10.1080/01621459.2026.2692052},
  note    = {arXiv:2210.08589}
}

@misc{grunwald2025supermartingales,
  title         = {Supermartingales for One-Sided Tests: Sufficient Monotone Likelihood Ratios are Sufficient},
  author        = {Gr{\"u}nwald, Peter D. and Koolen, Wouter M.},
  year          = {2025},
  eprint        = {2502.04208},
  archivePrefix = {arXiv},
  primaryClass  = {math.ST},
  url           = {https://arxiv.org/abs/2502.04208}
}

@article{goeman2011multiple,
  author  = {Goeman, Jelle J. and Solari, Aldo},
  title   = {Multiple Testing for Exploratory Research},
  journal = {Statistical Science},
  year    = {2011},
  volume  = {26},
  number  = {4},
  pages   = {584--597},
  doi     = {10.1214/11-STS356}
}

@article{katsevich2020simultaneous,
  author  = {Katsevich, Eugene and Ramdas, Aaditya},
  title   = {Simultaneous High-Probability Bounds on the False Discovery Proportion in Structured, Regression and Online Settings},
  journal = {The Annals of Statistics},
  year    = {2020},
  volume  = {48},
  number  = {6},
  pages   = {3465--3487},
  doi     = {10.1214/19-AOS1938}
}

@article{vovk2023confidence,
  author  = {Vovk, Vladimir and Wang, Ruodu},
  title   = {Confidence and Discoveries with {e}-Values},
  journal = {Statistical Science},
  year    = {2023},
  volume  = {38},
  number  = {2},
  pages   = {329--354},
  doi     = {10.1214/22-STS874}
}

@article{vovk2024truefalse,
  author  = {Vovk, Vladimir and Wang, Ruodu},
  title   = {True and False Discoveries with Independent and Sequential {e}-Values},
  journal = {The Canadian Journal of Statistics},
  year    = {2024},
  volume  = {52},
  number  = {4},
  pages   = {e11833},
  doi     = {10.1002/cjs.11833}
}

@article{vovk2025conformal,
  author  = {Vovk, Vladimir},
  title   = {Conformal {e}-Prediction},
  journal = {Pattern Recognition},
  year    = {2025},
  volume  = {166},
  pages   = {111674},
  doi     = {10.1016/j.patcog.2025.111674}
}

@misc{vovk2026confounding,
  author        = {Vovk, Vladimir and Wang, Ruodu},
  title         = {Conformal {e}-Prediction in the Presence of Confounding},
  year          = {2026},
  eprint        = {2603.11134},
  archivePrefix = {arXiv},
  primaryClass  = {math.ST},
  url           = {https://arxiv.org/abs/2603.11134}
}

@article{xu2024postselection,
  author  = {Xu, Ziyu and Wang, Ruodu and Ramdas, Aaditya},
  title   = {Post-selection Inference for {e}-Value Based Confidence Intervals},
  journal = {Electronic Journal of Statistics},
  year    = {2024},
  volume  = {18},
  number  = {1},
  pages   = {2292--2338},
  doi     = {10.1214/24-EJS2253}
}

\newpage

\section*{Appendix}\label{sec:appendix}

\section*{Related Work}\label{app:related_work}

\paragraph{Null proportion estimation.}
Our work builds on the literature of estimating the proportion of true null hypotheses in multiple testing. \cite{storey2002direct} introduced an estimator based on the proportion of large $p$-values, while \cite{meinshausen2006estimating} developed confidence bounds on the proportion of false null hypotheses. Related empirical Bayes approaches are discussed by \cite{efron2008microarrays}. These methods generally concern a prespecified collection of tests with $p$-values observed at a fixed time. In contrast, our setting allows decisions and their correctness labels to be generated adaptively, evidence collection to terminate at a data-dependent time, and the weights defining the target quantity to depend on the observed trajectory. We further characterize sharp bounds on expected agent quality over joint distributions of correctness labels and observed evidence compatible with a given observable distribution.

\paragraph{Sequential inference with $e$-values.}
The use of $e$-values is motivated by their suitability for inference under optional stopping and dependence \cite{vovk2021evalues,ramdas2023gametheoretic,grunwald2024safe}. Conditional $e$-values can be accumulated over time by predictable betting strategies, while arithmetic averages preserve validity across arbitrarily dependent $e$-values. \cite{vovk2024mergingsequentialevaluesmartingales} study martingale-based merging of sequential $e$-values, and \cite{wang2025admissiblewaymergingarbitrary} characterize admissible merging rules under arbitrary dependence. Our approach combines evidence both within and across hypotheses, with predictable weights for estimation and a shared wealth budget for confidence set construction. Furthermore, the sharpness result for our estimator class follows from the conditional moment restrictions on evidence evaluated at random correctness labels, rather than from an assumed model for their joint distribution.

\paragraph{Confidence bounds in multiple testing.}
Our confidence sets are related to simultaneous inference on the numbers of true and false hypotheses. \cite{goeman2011multiple} used closed testing to obtain confidence statements that remain valid after selecting a set of hypotheses, while \cite{katsevich2020simultaneous} developed simultaneous false discovery proportion bounds in structured and online settings. In the $e$-value literature, \cite{vovk2023confidence,vovk2024truefalse} construct confidence bounds for discovery counts, with the latter exploiting independence or sequential validity across the base $e$-values. \cite{xu2024postselection} obtain false coverage rate control for $e$-value-based confidence intervals reported after data-dependent selection. Our construction likewise inverts $e$-value tests, but targets a possibly data-dependent weighted sum of correctness indicators, with a potentially random number of hypotheses. Its shared betting process is designed to preserve validity when the true vector of correctness labels is itself random and may depend on the observed trajectory, unlike the approaches discussed above.

\paragraph{Predictive inference for random labels.}
Our inference approach is most closely related to conformal e-prediction \cite{vovk2025conformal}, which constructs e-values for candidate labels and guarantees validity at the random true label under exchangeability. Its conditional formulation imposes calibration restrictions involving label indicators, resembling our conditional moment constraints for random correctness labels. Related work by \cite{vovk2026confounding} employs a shared allocation across possible labels to ensure predictive validity in the presence of confounding. This is analogous to our shared betting weights, although their allocation is static whereas ours evolves predictably with sequential evidence. Unlike these predictive inference settings, our goal is to construct confidence sets for a data-dependent weighted sum of unobserved correctness labels associated with adaptively generated decisions. Our framework allows arbitrary dependence between labels and observed evidence, data-dependent stopping, and quality measures defined after the trajectory concludes. To accommodate these features, our construction uses a common predictable betting policy across candidate correctness vectors, with wealth transfers across hypotheses that preserve validity without requiring exchangeability.

\newpage

\section*{Lower Bound Estimation}\label{app:lower_bound}

For weights satisfying Assumption \ref{assump:integrable_pred_weights}, define
$$\widehat{\theta}_L^* =\sum_{i=1}^M \min\{c_i+A_i,\bar A_i\}.$$

We obtain the following analogues of Theorems \ref{thm:admiss_est} and \ref{thm:oracle_sharpness}. 

\begin{theorem}[Lower Bound Estimation]
Under Assumptions~1--5, $\widehat{\theta}_L^*$ is a uniformly valid lower bound estimator. Furthermore, let 
$$
L(\widetilde Q)
=
\inf_{Q\in\Pi(\widetilde Q)}\mathbb E_Q[\theta^*],
$$
for every $\tilde{Q}\in\tilde{\Qcal}$ and $\epsilon>0$, there exist weights satisfying Assumption~5 such that
\[
\mathbb E_{\widetilde Q}[\widehat{\theta}_L^*]
\ge L(\widetilde Q)-\epsilon.
\]
\end{theorem}

\begin{proof}
Replace $\theta_i$ by $1-\theta_i$ and interchange
$(e_{i,t},\bar e_{i,t})$ and $(w_{i,t},\bar w_{i,t})$.
The transformed problem satisfies the same assumptions,
with target $\sum_{i=1}^M c_i-\theta^*$ and upper
bound estimator
\[
\sum_{i=1}^M \max\{c_i-\bar A_i,-A_i\}
=
\sum_{i=1}^M c_i-\widehat{\theta}_L^*.
\]
Both conclusions therefore follow from
Theorems~1--2.
\end{proof}

\newpage
\section*{Proofs of Main Results}

\paragraph{Extended Conventions} For all $(w_{i,t}, \bar{w}_{i,t})$ and $(e_{i,t}, \bar{e}_{i,t})$ with index $(i,t)$ with $i > M$, $t > N$, or $R_{i,t}=0$, we set $(w_{i,t}, \bar{w}_{i,t}) = (0,0)$ and $(e_{i,t}, \bar{e}_{i,t}) = (1,1)$. For $R_{i,t}$ with $i > M$ or $t > N$, we set $R_{i,t} = 0$. For all $i > M$, we set $\theta_i = 0$ and $c_i = 0$. 

Before providing the proof for Theorems \ref{thm:admiss_est} and \ref{thm:oracle_sharpness}, we begin with a preliminary lemma that ensures the expectation of our estimator $\hat\theta_U^*$ is well defined. 

\begin{lemma}[Finite Expectation]\label{lem:finite_expectation}
    Under Assumptions \ref{assump:finite_expected_changes}-\ref{assump:integrable_pred_weights},  the expectation of $\hat\theta_*^U$ is finite with respect to all $Q \in \Qcal$.
\end{lemma}

\begin{proof}[Proof of Lemma \ref{lem:finite_expectation}]
    To prove Lemma \ref{lem:finite_expectation}, we first show that for all $Q \in \Qcal$.
    \begin{equation}
        {\EE_{Q}\left[ \mathbf{1}[\hat\theta_*^U > 0]|\hat\theta_*^U|\right]}  < \infty, \qquad  {\EE_{Q}\left[ \mathbf{1}[\hat\theta_*^U < 0]|\hat\theta_*^U|\right]} < \infty.
    \end{equation}
     By the nonnegativity of $(e_{i,t}, \bar{e}_{i,t})$ and $(w_{i,t}, \bar{w}_{i,t})$, 
    \begin{equation}
        -A_{i,} \leq \sum_{t =1}^\infty R_{i,t} w_{i,t}, \qquad -\bar{A}_i \leq \sum_{t=1}^\infty R_{i,t} \bar{w}_{i,t}. 
    \end{equation}
    Thus, when $\hat\theta_*^U > 0$, the absolute value of estimator is pathwise bounded by 
    \begin{align}
        \left|\hat\theta_*^U\right| &= \left|\sum_{i=1}^\infty \mathbf{1}[i \leq M]\max\{c_i - A_i, -\bar{A}_i\}\right|\\
        &\leq \sum_{i=1}^\infty \mathbf{1}[i \leq M]\max\left\{ c_i + \sum_{t =1}^\infty R_{i,t}w_{i,t}, \sum_{t=1}^\infty R_{i,t}\bar{w}_{i,t} \right\}\\
        &\leq \sum_{i=1}^\infty \mathbf{1}[i \leq M] \left[ c_i + \sum_{t=1}^\infty R_{i,t}(w_{i,t} + \bar{w}_{i,t})   \right]\\
        &= \sum_{i=1}^\infty \mathbf{1}[i \leq M] c_i + \sum_{i=1}^\infty \mathbf{1}[i \leq M]\sum_{t=1}^\infty \mathbf{1}[t \leq N] R_{i,t}(w_{i,t} + \bar{w}_{i,t}).
    \end{align}
    Therefore, we obtain for all $Q \in \Qcal$,
    \begin{equation}
        \EE_{Q}\left[ \mathbf{1}[\hat\theta_*^U > 0]|\hat\theta_*^U|\right] \leq \EE_Q\left[ \sum_{i=1}^M c_i \right] + \EE_{Q}\left[ \sum_{i=1}^M \sum_{t=1}^N R_{i,t}(w_{i,t} + \bar{w}_{i,t}) \right] < \infty,
    \end{equation}
    by Assumptions \ref{assump:target_param_restrictions}--\ref{assump:integrable_pred_weights}. On the event where $\hat\theta_*^U < 0$, we have $|\hat\theta_*^U| = - \hat\theta_*^U$. 
    
    By $\theta_i \in \{0,1\}$ and $w_{i,t}, \bar{w}_{i,t}\geq 0$, 
    \begin{align}
        |\hat\theta_*^U| &= \sum_{i=1}^\infty \mathbf{1}[i \leq M] \min\{A_i - c_i, \bar{A}_i\} \\
        &\leq \sum_{i=1}^\infty  \mathbf{1}[i \leq M] [\theta_i (A_i - c_i) + (1-\theta_i)\bar{A}_i] \\
        &\leq \sum_{i=1}^\infty \sum_{t=1}^\infty R_{i,t}[\theta_i w_{i,t}e_{i,t} + (1-\theta_i)\bar{w}_{i,t} \bar{e}_{i,t}], \label{eq:1}
    \end{align}
    where the final term is nonnegative by the fact that $R_{i,t}, \theta_i \in \{0,1\}$, the nonnegativity of $(e_{i,t}, \bar{e}_{i,t})$, and the nonnegativity of $(w_{i,t}, \bar{w}_{i,t})$. Furthermore, by Assumption \ref{assump:e_value_constraint},  
    \begin{equation}
        \EE_Q\left[ R_{i,t} \theta_i w_{i,t}  e_{i,t} \mid \Fcal_{t-1}\right] \leq R_{i,t}w_{i,t} , \qquad \EE_{Q}\left[R_{i,t} (1-\theta_i) \bar{w}_{i,t} \bar{e}_{i,t} \mid \Fcal_{t-1}\right]  \leq R_{i,t} \bar{w}_{i,t}. \label{eq:2} 
    \end{equation}
    for all $Q \in \Qcal$. Combining the results of Equation \eqref{eq:1}--\eqref{eq:2}, we obtain by Tower's law, 
    \begin{align}
        \EE_{Q}\left[ \mathbf{1}[\hat\theta_*^U < 0]|\hat\theta_*^U|\right] &\leq \EE_{Q}\left[\sum_{i=1}^\infty \sum_{t=1}^\infty R_{i,t}[\theta_i w_{i,t}e_{i,t} + (1-\theta_i)\bar{w}_{i,t} \bar{e}_{i,t}]\right] \label{eq:ref_1}\\
        &\leq \EE_{Q}\left[ \sum_{i=1}^\infty \sum_{t=1}^\infty \EE_{Q}\left[R_{i,t}[\theta_i w_{i,t} e_{i,t} + (1-\theta_i)\bar{w}_{i,t} \bar{e}_{i,t}] \mid \Fcal_{t-1} \right] \right] \\
        &\leq \EE_{Q}\left[ \sum_{i=1}^\infty \mathbf{1}[i \leq M] \sum_{t=1}^\infty \mathbf{1}[t \leq N] R_{i,t}(w_{i,t} + \bar{w}_{i,t}) \right]\\
        &= \EE_{Q}\left[ \sum_{i=1}^M \sum_{t=1}^N R_{i,t}(w_{i,t} + \bar{w}_{i,t}) \right]\\
        &< \infty, \label{eq:ref_2}
    \end{align}
    where the third line follows from the fact that for all $(i,t)$ where $i >M$ or $t > N$, the inner summand is zero, and the last line holds by Assumption \ref{assump:integrable_pred_weights}. Then, 
    \begin{equation}
        \EE_{Q}\left[ \hat\theta_*^U \right] = \underbrace{\EE_{Q}\left[ \mathbf{1}[\hat\theta_*^U > 0]|\hat\theta_*^U|\right]}_{< \infty} - \underbrace{\EE_{Q}\left[ \mathbf{1}[\hat\theta_*^U < 0]|\hat\theta_*^U|\right]}_{< \infty} < \infty,
    \end{equation}
    so the expectation of $\hat\theta_*^U$ is well-defined and finite.
\end{proof}

\subsection*{Proof of Theorem \ref{thm:admiss_est}}

We first show that the expectation $\hat\theta_*^U - \theta_*$ is finite. By Lemma \ref{lem:finite_expectation}, $\EE_{Q}\left[\hat\theta_*^U\right]$ is finite for all $Q \in \Qcal$. Furthermore, by the nonnegativity of $c_i \geq 0$ and $\theta_i \in \{0,1\}$, we obtain $\sum_{i=1}^M\ c_i \theta_i \leq  \sum_{i=1}^M c_i$. By Assumption \ref{assump:target_param_restrictions} and the previous inequality, for all $Q \in \Qcal$,  
\begin{equation}
    \EE_{Q}\left[ \theta_* \right] \leq \EE_{Q}\left[ \sum_{i=1}^M c_i \right] < \infty.
\end{equation}
Therefore, $\EE_{Q}\left[\hat\theta_*^U - \theta_* \right]$ is well-defined, and it suffices to show that $\EE_{Q}\left[ \hat\theta_*^U - \theta_* \right] \geq 0$. 

We start by constructing pathwise lower bounds for $(\hat\theta_*^U - \theta_*)$. Rewriting the difference, 
\begin{align}
    \hat\theta_*^U - \theta_* &= \sum_{i=1}^\infty \mathbf{1}[i \leq M] (\psi_i - c_i \theta_i). \label{eq:3}
\end{align}
By definition, the term $\psi_i$ can be lower bounded as 
\begin{equation}
    \psi_i  = \max\left( c_i - A_i , -\bar{A}_i \right) \geq \theta_i(c_i - A_i) - (1-\theta_i) \bar{A}_i, \label{eq:4}
\end{equation}
and therefore by Equations \eqref{eq:3}--\eqref{eq:4}, we obtain the lower bound
\begin{equation}
     \hat\theta_*^U - \theta_* \geq -\sum_{i=1}^\infty\mathbf{1}[i \leq M](\theta_i A_i + (1-\theta_i)\bar{A}_i). 
\end{equation}
By expanding the terms $A_i$ and $\bar{A}_i$, we obtain
\begin{align}
    \hat\theta_*^U - \theta_* &\geq -\sum_{i=1}^\infty\mathbf{1}[i \leq M]\left(\sum_{t=1}^\infty \mathbf{1}[t \leq N] R_{i,t}\left[\theta_i w_{i,t} (e_{i,t}-1) + (1-\theta_i)\bar{w}_{i,t}(\bar{e}_{i,t}-1)\right]  \right)\\
    &= -\sum_{i=1}^\infty \sum_{t=1}^\infty R_{i,t}\left[\theta_i w_{i,t} (e_{i,t}-1) + (1-\theta_i)\bar{w}_{i,t}(\bar{e}_{i,t}-1)\right]  \label{eq:estimator-pathwise-lower-bound},
\end{align}
where the last line follows from $R_{i,t} = 0$ for all $t > N$ or $i > M$. Before applying expectations with respect to $Q \in \Qcal$, note that $\sum_{i=1}^\infty \sum_{t=1}^\infty R_{i,t}\left[\theta_i w_{i,t} e_{i,t} + (1-\theta_i)\bar{w}_{i,t}\bar{e}_{i,t}\right]$ is integrable by lines \eqref{eq:ref_1}--\eqref{eq:ref_2} in the proof of Lemma \ref{lem:finite_expectation}. Furthermore, note that $\sum_{i=1}^\infty \sum_{t=1}^\infty R_{i,t}(\theta_iw_{i,t} + (1-{\theta_i})\bar{w}_{i,t}) \leq \sum_{i=1}^M\sum_{t=1}^N R_{i,t} (w_{i,t} + \bar{w}_{i,t}) $, and therefore is integrable by Assumption \ref{assump:integrable_pred_weights}. Thus,
\begin{align}
    &-\sum_{i=1}^\infty \sum_{t=1}^\infty R_{i,t}\left[\theta_i w_{i,t} (e_{i,t}-1) + (1-\theta_i)\bar{w}_{i,t}(\bar{e}_{i,t}-1)\right] = \\
    &\sum_{i=1}^\infty \sum_{t=1}^\infty R_{i,t}(\theta_iw_{i,t} + (1-{\theta_i})\bar{w}_{i,t}) -\sum_{i=1}^\infty \sum_{t=1}^\infty R_{i,t}\left[\theta_i w_{i,t} e_{i,t} + (1-\theta_i)\bar{w}_{i,t}\bar{e}_{i,t}\right]
\end{align}
is also integrable, as it is the sum of two integrable functions. Then, by Fubini's Theorem
\begin{align}
   \EE_Q\left[ \hat\theta_*^U - \theta_* \right] &\geq -   \EE_{Q}\left[ \sum_{i=1}^\infty \sum_{t=1}^\infty R_{i,t}\left[\theta_i w_{i,t} (e_{i,t}-1) + (1-\theta_i)\bar{w}_{i,t}(\bar{e}_{i,t}-1)\right]  \right]\\
    &= \sum_{i=1}^\infty \sum_{t=1}^\infty \EE_{Q}\left[R_{i,t}\left[\theta_i w_{i,t} (1-e_{i,t}) + (1-\theta_i)\bar{w}_{i,t}(1-\bar{e}_{i,t})\right]  \right]\label{eq:5}
\end{align}
Note that conditional on $\Fcal_{t-1}$, by Assumption \ref{assump:e_value_constraint}, for all $(i,t) \in \NN \times \NN$, 
\begin{align}
    &\EE_{Q}\left[R_{i,t}\left[\theta_i w_{i,t} (1-e_{i,t}) + (1-\theta_i)\bar{w}_{i,t}(1-\bar{e}_{i,t})\right]  \right]\\
    &= \EE_{Q}\left[ \EE_{Q}\left[ \theta_i R_{i,t} w_{i,t}(1-e_{i,t}) + (1-\theta_i)R_{i,t}\bar{w}_{i,t}(1-\bar{e}_{i,t}) \mid \Fcal_{t-1}\right] \right] \\
    &= \EE_{Q}\left[ w_{i,t}\EE_{Q}\left[ \theta_iR_{i,t}(1-e_{i,t}) \mid \Fcal_{t-1}] +  \bar{w}_{i,t}\EE_{Q}[(1-\theta_i)R_{i,t}(1-\bar{e}_{i,t}) \mid \Fcal_{t-1}\right] \right]\\
    &\geq 0\label{eq:6},
\end{align}
where the last line follows from Assumption \ref{assump:e_value_constraint} and Assumption \ref{assump:integrable_pred_weights}. By Equations \eqref{eq:5}-- \eqref{eq:6}, 
\begin{equation}
    \EE_Q\left[ \hat\theta_*^U - \theta_* \right] \geq \sum_{i=1}^\infty \sum_{t=1}^\infty \EE_{Q}\left[R_{i,t}\left[\theta_i w_{i,t} (1-e_{i,t}) + (1-\theta_i)\bar{w}_{i,t}(1-\bar{e}_{i,t})\right]  \right] \geq 0,
\end{equation}
obtaining our desired result for any $Q \in \Qcal$. Because this holds for all $Q \in \Qcal$, we obtain our desired result.

\subsection*{Proof of Theorem \ref{thm:oracle_sharpness}}

Our proof proceeds in four steps. First, for a fixed $\tilde{Q} \in \tilde{\Qcal}$, we characterize the distributions $Q \in \Pi(\tilde{Q})$ through terminal correctness probabilities, and show that our reformulation has the same attainable expected target quantity $\theta_*$. Second, using our terminal correctness probability characterization, we represent feasibility by taking the Lagrangian for our $e$-value moment constraints, obtaining predictable multipliers. Third, we swap the minimization and maximization operations in the dual via Sion's Minimax Theorem. Lastly, by evaluating the inner supremum, we establish our desired sharpness result.

\paragraph{Key Conditions.}
Our proofs primarily rely on Assumptions \ref{assump:e_value_constraint} and \ref{assump:target_param_restrictions}. Assumption \ref{assump:e_value_constraint} requires nonnegative $e$-values, the integrability condition
\begin{equation}\label{eq:t2-assumption-integrability}
\mathbb E_Q\!\left[\sum_{i=1}^M\sum_{t=1}^N
 R_{i,t}(1+e_{i,t}+\bar e_{i,t})\right]<\infty,
\end{equation}
together with the conditional moment inequalities
\begin{equation}\label{eq:t2-assumption-moments}
\EE_Q[\theta_i R_{i,t}(e_{i,t}-1)\mid\mathcal F_{t-1}]\leq0,
\qquad
\EE_Q[(1-\theta_i)R_{i,t}(\bar e_{i,t}-1)\mid\mathcal F_{t-1}]\leq0.
\end{equation}
The integrability condition in Equation \eqref{eq:t2-assumption-integrability}, nonnegative $e$-values, and $R_{i,t} \in \{0,1\}$ yield
\begin{equation}\label{eq:t2-increments}
\begin{aligned}
D&:=\mathbb E_{\tilde Q}\!\left[\sum_{i,t\ge1}R_{i,t}(|e_{i,t}-1|+|\bar e_{i,t}-1|)\right]\\
&\le\mathbb E_{\tilde Q}\!\left[\sum_{i=1}^M\sum_{t=1}^N
R_{i,t}(2+e_{i,t}+\bar e_{i,t})\right]<\infty.
\end{aligned}
\end{equation}

Under Assumption \ref{assump:target_param_restrictions}, $c_i\geq 0$, $c_i$ is $\Fcal_N$ measurable, and $\mathbb E_{\tilde Q}[\sum_i c_i]<\infty$.

\subsubsection*{Step 1: Equivalence Characterization with Terminal Probabilities.}
We begin by fixing $\tilde{Q} \in \tilde{\Qcal}$. Let
$$
\begin{gathered}
\mathcal X=\{(p_i)_{i\ge1}:p_i\text{ is }\mathcal F_N\text{-measurable and }0\le p_i\le \mathbf1\{i\le M\}\},\\
J(p)=\mathbb E_{\tilde Q}\!\left[\sum_{i\ge1}c_i p_i\right].
\end{gathered}
$$
Let $\mathcal X_{\mathrm{feas}}\subseteq\mathcal X$ be the subset satisfying, for every $i,t\ge1$,
\begin{equation}\label{eq:t2-observable-moments}
\begin{aligned}
\mathbb E_{\tilde Q}[p_iR_{i,t}(e_{i,t}-1)\mid\mathcal F_{t-1}]&\le0,\\
\mathbb E_{\tilde Q}[(1-p_i)R_{i,t}(\bar e_{i,t}-1)\mid\mathcal F_{t-1}]&\le0.
\end{aligned}
\end{equation}
Our objective in this step is the exact reformulation
\begin{equation}\label{eq:t2-primal}
U(\tilde Q)=\sup_{Q\in\Pi(\tilde Q)}\mathbb E_Q[\theta_*]
=\sup_{p\in\mathcal X_{\mathrm{feas}}}J(p).
\end{equation}
We establish it by proving the forward construction, the reverse construction, and preservation of the objective under both constructions.

\textit{Compatible Joint Distribution to Terminal Probabilities.}
For $Q\in\Pi(\tilde Q)$, let $$p_i^Q=\mathbb E_Q[\theta_i\mid\mathcal F_N].$$ Since $\mathbf1\{i\leq M\}$ is $\Fcal_N$-measurable and $0\le\theta_i\le \mathbf1\{i\le M\}$, this gives $p^Q\in\mathcal X$. Furthermore, since $R_{i,t}(e_{i,t}-1)$ is $\Fcal_N$-measurable and $\Fcal_{t-1} \subseteq \Fcal_N$, 
\begin{align}
    \EE_{Q}[\theta_iR_{i,t}(e_{i,t} -1) \mid \Fcal_{t-1}] &= \EE_{Q}\left[\EE_{Q}\left[ \theta_i R_{i,t}(e_{i,t}-1) \mid \Fcal_N \right] \mid \Fcal_{t-1} \right] \\
    &= \EE_{\tilde{Q}}\left[ R_{i,t}(e_{i,t}-1)\EE_{Q}\left[\theta_i \mid \Fcal_N \right] \mid \Fcal_{t-1}\right]\\
    &= \EE_{\tilde{Q}}\left[ R_{i,t}(e_{i,t}-1)p_i^Q \mid \Fcal_{t-1}\right]\\
    &\leq 0,
\end{align}
by Assumption \ref{assump:e_value_constraint}. By repeating our step above for $\bar{e}_{i,t}$ and $(1-\theta_i)$, we obtain 
\begin{equation}\label{eq:t2-project}
    \EE_{\tilde{Q}}\left[ p_i^Q R_{i,t} (e_{i,t}-1)\mid \Fcal_{t-1} \right] \leq 0, \qquad \EE_{\tilde{Q}}\left[ \left(1-p_i^Q\right) R_{i,t} (\bar{e}_{i,t}-1) \mid \Fcal_{t-1}\right] \leq 0.
\end{equation}
Thus, $p^Q$ belongs to our feasible set $\Xcal_{\mathrm{feas}}$.

\emph{From feasible terminal probabilities to a compatible joint law.} Conversely, fix any $p\in\Xcal_{\mathrm{feas}}$. Construct a joint distribution $Q^p$ with observable marginal $\tilde{Q}$ by specifying the conditional distribution of the
correctness labels as
$$
    \theta_i\mid\mathcal F_N
    \sim \operatorname{Bernoulli}(p_i),
    \qquad i\in\mathbb N,
$$
independently across $i$. By construction,
$\mathbb E_{Q^p}[\theta_i\mid\mathcal F_N]=p_i$.
Thus, for every $i,t \geq 1$, by $\Fcal_{t-1} \subseteq \Fcal_N$
\begin{align*}
\mathbb E_{Q^p}
[\theta_iR_{i,t}(e_{i,t}-1)\mid\mathcal F_{t-1}]=
\mathbb E_{\tilde Q}
[p_iR_{i,t}(e_{i,t}-1)\mid\mathcal F_{t-1}]
\le 0,
\end{align*}
where the inequality follows from
$p\in\mathcal X_{\mathrm{feas}}$.
The corresponding inequality for
$(1-\theta_i)R_{i,t}(\bar e_{i,t}-1)$
follows analogously. All remaining conditions in
Assumptions \ref{assump:finite_expected_changes}---\ref{assump:target_param_restrictions} are preserved, since $Q^p$ has the
same observable marginal as $\tilde Q$ and
$\theta_i\in\{0,1\}$, with $\theta_i=0$ for $i>M$.
Therefore, $Q^p\in\Pi(\tilde Q)$.


\emph{Preservation of Objective.}
Because $c_i$ is $\Fcal_N$-measurable and its sum has finite expectation under Assumption \ref{assump:target_param_restrictions}, we obtain 
\[
\mathbb E_Q[\theta_*]
=\mathbb E_{\tilde Q}\!\left[\sum_i c_i\mathbb E_Q[\theta_i\mid\mathcal F_N]\right]
=J(p^Q).
\]
The reverse construction similarly gives $\mathbb E_{Q^p}[\theta_*]=J(p)$. Hence the two feasible sets produce exactly the same attainable expected target values, which proves \eqref{eq:t2-primal}.

\paragraph{Step 2: Lagrangian Dual with Predictable Multipliers}
We take the dual of our objective with respect to $\Xcal$. Let $\Wcal_b$ denote the class of nonnegative, $\Fcal_{t-1}$-measurable weight sequences $(w_{i,t}, \bar{w}_{i,t})_{i,t \geq 1}$ that equal to zero whenever $R_{i,t} = 0$ and for which there exists a finite constant $C < \infty$ such that $0 \leq w_{i,t}, \bar{w}_{i,t} \leq C$ for all $i,t$. The conditions above hold pointwise for $\Wcal_b$. Note that the upper bound $C$ can depend on the sequence, and is not fixed across the class. 

First, for $(w,\bar w)\in\mathcal W_b$, define $A_i$ and $\bar{A}_i$ as in Equation \eqref{eq:e_processes},
$$
A_i=\sum_{t\ge1}w_{i,t}R_{i,t}(e_{i,t}-1),\qquad
\bar A_i=\sum_{t\ge1}\bar w_{i,t}R_{i,t}(\bar e_{i,t}-1),
$$
The Lagrangian can then be written as
\begin{align}
\Lcal(p,w,\bar{w})
&=J(p)-\sum_{i,t\geq1}\EE_{\tilde Q}\!\left[
 R_{i,t}\{w_{i,t}p_i(e_{i,t}-1)+\bar w_{i,t}(1-p_i)(\bar e_{i,t}-1)\}\right] \label{eq:first_line}\\
&= \EE_{\tilde{Q}}\left[ \sum_{i \geq 1} c_ip_i \right] -\sum_{i,t\geq1}\EE_{\tilde Q}\!\left[
 R_{i,t}\{w_{i,t}p_i(e_{i,t}-1)+\bar w_{i,t}(1-p_i)(\bar e_{i,t}-1)\}\right] \\
&=\mathbb E_{\tilde Q}\!\left[\sum_{i\geq1}
 \{p_i(c_i-A_i)-(1-p_i)\bar A_i\}\right].
\label{eq:t2-lagrangian}
\end{align}
Note that our Lagrangian expression is finite under Assumption \ref{assump:target_param_restrictions} and Equation \eqref{eq:t2-increments} for any $p \in \Xcal$ and $(w, \bar{w}) \in \Wcal_b$, allowing for the interchange of expectation and summation operators above. To show this result, for weight bound $C < \infty$,
\begin{align*}
    \left|\Lcal(p,w,\bar{w})\right| &\leq \EE_{\tilde{Q}}\left[\sum_{i\geq1}
 \left|p_i(c_i-A_i)-(1-p_i)\bar A_i\right| \right]\\
 & \leq \EE_{\tilde{Q}}\left[ \sum_{i \geq 1 } \left(c_i + |A_i| + |\bar{A}_i|\right)  \right]\\
 &\leq \EE_{\tilde{Q}}\left[\sum_{i=1}^M c_i \right] + C\EE_{\tilde{Q}}\left[ \sum_{i, t \geq 1} R_{i,t}(|e_{i,t}-1| + |\bar{e}_{i,t}-1|) \right]\\
 &\leq \EE_{\tilde{Q}}\left[\sum_{i=1}^M c_i \right] + CD < \infty, 
\end{align*}
where $D$ is defined as in Equation \eqref{eq:t2-increments}. 

We now establish
\begin{equation}\label{eq:t2-inner-inf}
\inf_{(w,\bar w)\in\mathcal W_b}\mathcal L(p,w,\bar w)
=\begin{cases}
J(p),&p\in\mathcal X_{\mathrm{feas}},\\
-\infty,&p\in\mathcal X\setminus\mathcal X_{\mathrm{feas}}.
\end{cases}
\end{equation}


First, assume $p\in\mathcal X_{\mathrm{feas}}$. Since weights are nonnegative and predictable, the conditional moment constraints for $\Xcal_{\mathrm{feas}}$ imply $\Lcal(p,w,\bar w)\ge J(p)$, with equality at $w_{i,t} = \bar{w}_{i,t} = 0$ for all $i,t$. Therefore $\mathcal L(p,w,\bar w)\ge J(p)$, with equality at zero weights. 

If $p\notin\mathcal X_{\mathrm{feas}}$, suppose the first moment restriction fails at $(i,t)$. Write $g=\mathbb E_{\tilde Q}[p_iR_{i,t}(e_{i,t}-1)\mid\mathcal F_{t-1}]$ and $B=\{g>0\}$, where $\tilde Q(B)>0$. Let $w_{i,t}=aR_{i,t}\mathbf1_B$, where $a>0$ is a fixed constant, and set all other multipliers to zero. Both $R_{i,t}$ and $\mathbf1_B$ are $\mathcal F_{t-1}$-measurable, so this policy lies in $\mathcal W_b$ for every finite $a>0$. Since $R_{i,t}^2=R_{i,t}$, sending $a \rightarrow \infty$ yields
$$J(p) - a\EE_{\tilde{Q}}[\mathbf{1}_B g] \longrightarrow -\infty .$$
A failure of the second restriction is treated using the analogous $\bar w_{i,t}$. This proves Equation \eqref{eq:t2-inner-inf}, and therefore the supremum over $p\in\mathcal X$ now yields
\begin{equation}\label{eq:t2-supinf}
U(\tilde Q)=\sup_{p\in\mathcal X}\inf_{(w,\bar w)\in\mathcal W_b}\mathcal L(p,w,\bar w).
\end{equation}

\paragraph{Step 3: Minimax Interchange.}
We now show
\begin{equation}\label{eq:t2-minimax}
\sup_{p\in\mathcal X}\inf_{(w,\bar w)\in\mathcal W_b}\mathcal L(p,w,\bar w)
=\inf_{(w,\bar w)\in\mathcal W_b}\sup_{p\in\mathcal X}\mathcal L(p,w,\bar w).
\end{equation}
To do so, we verify the hypotheses of Sion's minimax theorem (Corollary 3.3, \cite{sion1958general}).

\smallskip\noindent\emph{Compactness and convexity.}
Give each coordinate of $\mathcal X$ the weak-* topology $\sigma(L^\infty,L^1)$ for $(\mathcal F_N,\tilde Q)$. The set $\{p_i:0\le p_i\le\mathbf{1}\{i\le M\}\}$ is weak-* closed, since its order restrictions are equivalent to 
$$\EE_{\tilde{Q}}[hp_i] \geq 0, \qquad \EE_{\tilde{Q}}\left[ h(\mathbf{1}\{i \leq M\}) - p_i\right] \geq 0 $$
for every nonnegative $h \in L^1(\Fcal_N)$. It is also contained in the unit ball of $L^\infty(\mathcal F_N,\tilde Q)$.
By the Banach--Alaoglu theorem \citep{alaoglu1940weak},
each coordinate set is weak-* compact.
Tychonoff's theorem \citep{tychonoff1930topological}
implies that their infinite product $\Xcal$
is compact in the product topology. Thus, the product $\mathcal X$ is compact in the product of the weak-* topologies and is convex. We equip bounded predictable weight sequences $(w_{i,t}, \bar{w}_{i,t})$ with the pointwise supremum norm. Let $\Omega$ denote our sample space, with $\omega$ denoting a path. Then, 
$$
\|(w,\bar w)\|_\infty
=\sup_{i,t\geq1, \omega \in \Omega}\max\{|w_{i,t}(\omega)|,|\bar w_{i,t}(\omega)|\}.
$$
The set $\Wcal_b$ is convex in this normed linear space.

\smallskip\noindent\emph{Continuity in the terminal probabilities.}
Fix weight sequences $(w_{i,t}, \bar{w}_{i,t}) \in \Wcal_b$. Equation \eqref{eq:t2-lagrangian}
shows that $\Lcal$ is affine in $p$. Furthermore, by Assumption \ref{assump:target_param_restrictions} and Equation \eqref{eq:t2-increments}
\begin{equation}\label{eq:t2-tail}
\sum_{i\ge1}\mathbb E_{\tilde Q}[c_i+|A_i|+|\bar A_i|]
\le\mathbb E_{\tilde Q}\!\left[\sum_{i=1}^M c_i\right]+CD<\infty,
\end{equation}
where $C$ is the weight bound for $(w_{i,t}, \bar{w}_{i,t})$. Thus, the coefficient of $p_i$ in our Lagrangian in Equation \eqref{eq:t2-lagrangian} is the $L^1(\Fcal_N)$ function $c_i-A_i+\bar A_i$ for each $i$. Each finite partial sum is therefore weak-* continuous in its finitely many coordinates. 
Uniformly over $p\in\mathcal X$, its absolute expected tail after index $K$ is bounded by
$$ \sum_{i>K}\mathbb E_{\tilde Q}[c_i+|A_i|+|\bar A_i|],$$
which tends to zero. Therefore, the Lagrangian $\Lcal$ is continuous and affine in $p$ on $\mathcal X$.

\smallskip\noindent\emph{Continuity in the multipliers.}
For fixed $p$, Equation \eqref{eq:t2-lagrangian} also shows $\Lcal$ is affine in $(w, \bar{w})$. By Equation \eqref{eq:t2-increments}, 
\begin{equation}
    \left| L(p, w, \bar{w}) - L(p, w', \bar{w}' )  \right| \leq D\|(w, \bar{w}) - (w', \bar{w}') \|_{\infty}
\end{equation}
where $(w, \bar{w}) - (w', \bar{w}')$ denotes the pointwise differences in weight sequences. Therefore, $\Lcal$ is continuous and affine on $\Wcal_b$ with fixed choice of $p$. 

Applying Sion's theorem therefore proves our desired result in Equation \eqref{eq:t2-minimax}. Combining with the results of Equation \eqref{eq:t2-supinf}, we obtain 
\begin{equation}\label{eq:t2-dual}
U(\tilde Q)=\inf_{(w,\bar w)\in\mathcal W_b}\sup_{p\in\mathcal X}\mathcal L(p,w,\bar w).
\end{equation}

\medskip\noindent\textbf{Step 4: Inner Supremum Evaluation}
We first prove, for every fixed $(w,\bar w)\in\mathcal W_b$, 
\begin{equation}\label{eq:t2-inner-sup}
\sup_{p\in\mathcal X}\mathcal L(p,w,\bar w)
=\mathbb E_{\tilde Q}[\hat\theta_*^U],
\qquad
\hat\theta_*^U=\sum_{i=1}^M\max\{c_i-A_i,-\bar A_i\}.
\end{equation}
Let $\psi_i=\max\{c_i-A_i,-\bar A_i\}$. On $\{i\le M\}$, the $i$-th summand of the Lagrangian is
$p_i(c_i-A_i)+(1-p_i)(-\bar A_i)$. Its pointwise maximum with respect to $0\leq p_i\leq1$ is given by
\begin{equation}
    \sup_{0 \leq p \leq 1} \{p_i(c_i - A_i) - (1-p_i)\bar{A}_i \} = (c_i - A_i + \bar{A}_i)_+ - \bar{A}_i = \psi_i. 
\end{equation}
On $\{i>M\}$, by our appendix conventions, the summand is zero. The choice of sequence $ p_i^*=\mathbf1\{i\leq M,\ c_i-A_i\geq-\bar A_i\} $ is $\Fcal_N$-measurable, belongs to $\Xcal$, and attains these maxima. 

Since $p^* \in \Xcal$ attains the pointwise maximum for every $i$, and the expectation of $\Lcal$ is bounded by Equation \eqref{eq:t2-tail}, we obtain our desired result in Equation \eqref{eq:t2-inner-sup}. Substituting Equation \eqref{eq:t2-inner-sup} into Equation \eqref{eq:t2-dual} yields 
\begin{equation}\label{eq:t2-bounded}
U(\tilde Q)=\inf_{(w,\bar w)\in\mathcal W_b}\mathbb E_{\tilde Q}[\hat\theta_*^U].
\end{equation}

It remains to establish the same equality over the weights defined in Assumption \ref{assump:integrable_pred_weights}, which we denote as $\mathcal W_5$. Note that for any weights in $\Wcal_b$, 
\begin{equation}
    \EE_{Q}\left[ \sum_{i=1}^M \sum_{t=1}^N R_{i,t}(w_{i,t} + \bar{w}_{i,t}) \right] \leq 2C\EE_{Q}\left[ \sum_{i=1}^M \sum_{t=1}^N R_{i,t}\right] < \infty
\end{equation}
by the integrability constraint in Assumption \ref{assump:e_value_constraint}. Therefore, $\Wcal_b \subseteq \Wcal_5$, and Equation \eqref{eq:t2-bounded} and Theorem \ref{thm:admiss_est} yield
$$
U(\tilde{Q}) \leq \inf_{(w,\bar w)\in\mathcal W_5}\EE_{\tilde Q}[\hat\theta_*^U]
\leq\inf_{(w,\bar w)\in\mathcal W_b}\EE_{\tilde Q}[\hat\theta_*^U]=U(\tilde Q).
$$

Thus, Theorem \ref{thm:oracle_sharpness} holds by definition of infimum. In particular, for every $\epsilon>0$, there exists a bounded predictable policy with expected estimate at most $U(\tilde Q)+\epsilon$.

\subsection*{Proof of Lemma \ref{lem:domnating}}
We show that $\widehat{\theta}^{*}_{U} \leq
\pi^{\mathrm{Storey}}(\lambda)$ deterministically.
Under our choice of $\bar{w}_i = 0$ and
$w_i = \lambda/[M(1-\lambda)]$, the function
$\psi_i$ corresponds to
\begin{equation}
    \psi_i = \left(\frac{1}{M} -\frac{\lambda(e_i-1)}{M(1-\lambda)}\right)_{+} = \frac{(1-\lambda e_i)_{+}}{M(1-\lambda)}.
\end{equation}
We compare this term with the $i$-th summation
term in $\pi^{\mathrm{Storey}}(\lambda)$, denoted by
\begin{equation}
    \psi_i^{\mathrm{Storey}} = \frac{1}{M(1-\lambda)}
      \mathbf{1}\left[e_i \leq \frac{1}{\lambda}\right].
\end{equation}
When $e_i \leq 1/\lambda$, we deterministically obtain
\begin{equation}
    \psi_i = \frac{1-\lambda e_i}{M(1-\lambda)}
    \leq \frac{1}{M(1-\lambda)} = \psi_i^{\mathrm{Storey}},
\end{equation}
where the inequality follows from $e_i \geq 0$.
When $e_i > 1/\lambda$, both $\psi_i$ and
$\psi_i^{\mathrm{Storey}}$ are equal to zero.
Because this holds for every $i \in [M]$,
summing over hypotheses yields
\begin{equation}
    \widehat{\theta}^{*}_{U}
    = \sum_{i=1}^{M} \psi_i
    \leq \sum_{i=1}^{M} \psi_i^{\mathrm{Storey}}
    = \pi^{\mathrm{Storey}}(\lambda),
\end{equation}
establishing the desired pathwise domination.

\subsection*{Proof of Theorem \ref{thm:coverage}}

Fix any $Q \in \mathcal Q$, and let
$\theta=(\theta_i)_{i=1}^M$ denote the true
correctness labels. We first establish that
$K_t(\theta)$ is nonnegative and satisfies
$\mathbb E_Q[K_t(\theta)]\leq 1$ for every
$t\in\mathbb N$.

Our $e$-values are nonnegative and $\lambda_{i,t},\bar\lambda_{i,t}\in[0,1]$ by Assumptions \ref{assump:e_value_constraint} and \ref{assump:betting_and_transfers}, yielding
\begin{align}
A_{i,t} = A_{i,t-1}+R_{i,t}w_{i,t}(e_{i,t}-1) \geq A_{i,t-1}-w_{i,t}. 
\end{align}
By the constraint $w_{i,t} \leq D_{i, t-1} + b_{i,t}$, we also obtain 
\begin{equation}
    A_{i,t} \geq A_{i,t-1}-w_{i,t} \geq  A_{i,t-1}-D_{i,t-1}-b_{i,t} = \min\{A_{i,t-1},\bar A_{i,t-1}\}-b_{i,t}.
\end{equation}
The same inequality holds for $\bar A_{i,t}$, i.e. $\bar{A}_{i,t} \geq \min\{A_{i, t-1}, \bar{A}_{i,t-1}\}-b_{i,t}$. Then,
\begin{align}
B_t
&=1+\sum_{i=1}^M\min\{A_{i,t},\bar A_{i,t}\} \geq B_{t-1}-\sum_{i\in\mathcal A_t}b_{i,t}\geq 0,
\end{align}
where the last inequality follows 
from the budget constraint in
Assumption \ref{assump:betting_and_transfers}. Since
\begin{equation}
K_t(v)
=B_t+\sum_{i=1}^M
\big(v_iD_{i,t}+(1-v_i)\bar D_{i,t}\big)
\geq B_t,
\end{equation}
we obtain $K_t(\vartheta)\geq 0$ for every
$\vartheta\in\{0,1\}^M$, including $\vartheta=\theta$.

We now establish that
$\mathbb E_Q[K_t(\theta)]\leq 1$ for every
$t\in\mathbb N$. Define
\begin{align}
S_t=\sum_{i=1}^{\infty}R_{i,t}
\big(\theta_iw_{i,t}
+(1-\theta_i)\bar w_{i,t}\big), \qquad T_t
=\sum_{i=1}^{\infty}R_{i,t}
\big(\theta_iw_{i,t}e_{i,t}
+(1-\theta_i)\bar w_{i,t}\bar e_{i,t}\big).
\end{align}
Note that $K_t(\theta) = K_{t-1}(\theta) + T_t - S_t$ by definition. By Assumption \ref{assump:betting_and_transfers} and the definition of
$K_{t-1}(\theta)$ in Equation \eqref{eq:16}, 
\begin{align}\label{eq:useful_bound}
S_t \leq \sum_{i=1}^M
\big(\theta_iD_{i,t-1}
+(1-\theta_i)\bar D_{i,t-1}\big)
+\sum_{i\in\mathcal A_t}b_{i,t} \leq K_{t-1}(\theta).
\end{align}

We now showing $\mathbb E_Q[T_t]\leq
\mathbb E_Q[S_t]$ whenever
$\mathbb E_Q[K_{t-1}(\theta)]<\infty$.
For any $L>0$, let 
\begin{equation}
w_{i,t}^{(L)}=\min\{w_{i,t},L\},
\qquad
\bar w_{i,t}^{(L)}
=\min\{\bar w_{i,t},L\}.
\end{equation}
Let $S_t^{(L)}$ and $T_t^{(L)}$ denote
the corresponding sums obtained by replacing
$(w_{i,t},\bar w_{i,t})$ with
$(w_{i,t}^{(L)},\bar w_{i,t}^{(L)})$. By nonnegativity of the e-values and weights,
\begin{align}
0\leq T_t^{(L)}
&\leq L\sum_{i=1}^{\infty}
R_{i,t}(e_{i,t}+\bar e_{i,t}),\\
0\leq S_t^{(L)}
&\leq L\sum_{i=1}^{\infty}R_{i,t}
\leq LM.
\end{align}
Consequently, Assumptions \ref{assump:finite_expected_changes} and \ref{assump:e_value_constraint} imply
\begin{equation}
\EE_Q\left[T_t^{(L)}\right]<\infty,
\qquad
\mathbb E_Q[S_t^{(L)}]<\infty.
\end{equation}

Since the truncated weights are bounded,
nonnegative, and $\mathcal F_{t-1}$-measurable,
we apply Tonelli's theorem to our sum and the Tower
property within our expectation, yielding
\begin{align}
\mathbb E_Q[T_t^{(L)}]
&=\sum_{i=1}^{\infty}
\mathbb E_Q\Big[
w_{i,t}^{(L)}
\mathbb E_Q[
\theta_iR_{i,t}e_{i,t}
\mid\mathcal F_{t-1}]
\\
&\hspace{65pt}
+\bar w_{i,t}^{(L)}
\mathbb E_Q[
(1-\theta_i)R_{i,t}\bar e_{i,t}
\mid\mathcal F_{t-1}]
\Big]\\
&\leq\sum_{i=1}^{\infty}
\mathbb E_Q\Big[
w_{i,t}^{(L)}
\mathbb E_Q[
\theta_iR_{i,t}
\mid\mathcal F_{t-1}]
\\
&\hspace{65pt}
+\bar w_{i,t}^{(L)}
\mathbb E_Q[
(1-\theta_i)R_{i,t}
\mid\mathcal F_{t-1}]
\Big]\\
&=\mathbb E_Q[S_t^{(L)}],
\end{align}
where the inequality follows from the conditional
$e$-value constraints in Assumption \ref{assump:e_value_constraint}.

As $L\to\infty$, the nonnegative truncated sums $T_t^{(L)}$ and $S_t^{(L)}$ increase $Q$-almost-surely to $T_t$ and $S_t$ respectively. Therefore, by monotone convergence and Equation \eqref{eq:useful_bound},
\begin{align}\label{l:right_above}
\mathbb E_Q[T_t] =\lim_{L\to\infty}\mathbb E_Q[T_t^{(L)}] \leq\lim_{L\to\infty}\mathbb E_Q[S_t^{(L)}]=\mathbb E_Q[S_t] \leq\mathbb E_Q[K_{t-1}(\theta)]<\infty.
\end{align}
In particular, both $S_t$ and $T_t$ are
integrable whenever $K_{t-1}(\theta)$ is
integrable.

Since $K_0(\theta)=1$, we have
$\mathbb E_Q[K_0(\theta)]=1<\infty$.
Suppose that
$\mathbb E_Q[K_{t-1}(\theta)]\leq1$.
By Equation \eqref{l:right_above},
$\mathbb E_Q[T_t]\leq\mathbb E_Q[S_t]
\leq\mathbb E_Q[K_{t-1}(\theta)]<\infty$,
so all terms in the following expression are
integrable, and 
\begin{align}
\mathbb E_Q[K_t(\theta)]=\mathbb E_Q[K_{t-1}(\theta)]
-\mathbb E_Q[S_t]+\mathbb E_Q[T_t]\leq\mathbb E_Q[K_{t-1}(\theta)] \leq1.
\end{align}
By induction over $t \in \NN$, $\mathbb E_Q[K_t(\theta)]\leq1$
for every $t\in\mathbb N$.

By Assumption \ref{assump:finite_trajectory_length},
$N<\infty$ $Q$-almost surely.
Since $R_{i,t}=0$ for all $t>N$,
we have $K_t(\theta)=K_N(\theta)$ for
every $t\geq N$. Consequently,
$K_t(\theta)\to K_N(\theta)$
$Q$-almost surely as $t\to\infty$.
By nonnegativity of $K_t(\theta)$ and
Fatou's lemma,
\begin{equation}
\mathbb E_Q[K_N(\theta)]
\leq \liminf_{t\to\infty}
\mathbb E_Q[K_t(\theta)]
\leq1.
\end{equation}

Finally, recall that
$\theta^*=\sum_{i=1}^M c_i\theta_i$.
By the definition of $C(\alpha,F_N)$,
whenever $K_N(\theta)<1/\alpha$,
the true correctness vector $\theta$
belongs to $V(\alpha,F_N)$, implying
$\theta^*\in C(\alpha,F_N)$.
Therefore, 
\begin{align}
P_Q\big(\theta^*\notin C(\alpha,F_N)\big)
&\leq P_Q\big(K_N(\theta)\geq1/\alpha\big)\\
&\leq\alpha\mathbb E_Q[K_N(\theta)]\\
&\leq\alpha.
\end{align}
by Markov's inequality. Because this holds for every $Q\in\Qcal$,
we obtain
\begin{equation}
\sup_{Q\in\Qcal}
P_Q\big(\theta^*\notin C(\alpha,F_N)\big)
\leq\alpha,
\end{equation}
establishing the desired uniform coverage.

\newpage

\section{Empirical Tuning Procedures}\label{app:tuning}

The validity results in Sections~\ref{sec:estimator} and~\ref{sec:inference}
hold for every admissible predictable tuning rule. We therefore use conditional
trajectory simulation only to improve precision. At time $t$, let
$\hat Q_t(\cdot\mid\Fcal_{t-1})$ be an $\Fcal_{t-1}$-measurable simulator of
the remainder of the observed trajectory. All optimization and simulation at
time $t$ is completed before the time-$t$ $e$-values are observed, so the
resulting weights and transfers remain $\Fcal_{t-1}$-measurable.

\subsection{Empirical Tuning for Estimation}

For use below, define the running evidence terms
\begin{equation}\label{eq:est-running}
  A_{i,t}=\sum_{j=1}^{t}w_{i,j}R_{i,j}(e_{i,j}-1),\qquad
  \bar A_{i,t}=\sum_{j=1}^{t}\bar w_{i,j}R_{i,j}(\bar e_{i,j}-1),
\end{equation}
with $A_{i,0}=\bar A_{i,0}=0$. Thus $A_{i,N}=A_i$ and
$\bar A_{i,N}=\bar A_i$ in Equation~\eqref{eq:e_processes}.

Fix an observable trajectory law $\tilde Q\in\tilde\Qcal$. The oracle
continuation policy at time $t$ minimizes the conditional expected terminal
upper-bound estimator:
\begin{equation}\label{eq:estimator-oracle}
  \inf_{(w_{i,j},\bar w_{i,j})_{i\geq1,\,j\geq t}}
  \EE_{\tilde Q}\!\left[
    \sum_{i=1}^{M}
    \max\{c_i-A_{i,N},-\bar A_{i,N}\}
    \;\middle|\;\Fcal_{t-1}
  \right],
\end{equation}
where the optimization is over nonnegative, predictable continuation weights
satisfying Assumption~\ref{assump:integrable_pred_weights}, and the evidence
accumulated through time $t-1$ is held fixed. Because the terminal estimator is
additive across decisions and its weight restrictions are decision-specific,
the continuation problem separates across $i$. For an active decision
$i\in\Acal_t:=\{i:R_{i,t}=1\}$, its oracle problem has the epigraph form
\begin{align}
  &\inf_{\substack{(w_{i,j},\bar w_{i,j})_{j\geq t},\;
                    z_i\in L^1(\Fcal_N,\tilde Q)}}
    \EE_{\tilde Q}[z_i\mid\Fcal_{t-1}]
    \label{eq:oracle_single_1}\\
  &\text{subject to}\quad
    z_i\geq c_i-A_{i,t-1}
      -\sum_{j=t}^{N}w_{i,j}R_{i,j}(e_{i,j}-1),\nonumber\\
  &\hphantom{\text{subject to}\quad}
    z_i\geq-\bar A_{i,t-1}
      -\sum_{j=t}^{N}\bar w_{i,j}R_{i,j}(\bar e_{i,j}-1),\nonumber\\
  &\hphantom{\text{subject to}\quad}
    w_{i,j},\bar w_{i,j}\geq0,\quad
    w_{i,j},\bar w_{i,j}\text{ are }\Fcal_{j-1}\text{-measurable},
    \quad j\geq t.
    \label{eq:oracle_single_2}
\end{align}
For any fixed continuation policy, the smallest feasible $z_i$ equals the
terminal summand $\max\{c_i-A_{i,N},-\bar A_{i,N}\}$ almost surely.

We approximate this oracle problem with $S$ simulated completions. For
$s\in[S]$, completion $s$ contains a terminal horizon $N^{(s)}$, a terminal
number of decisions $M^{(s)}$, observation indicators $R_{i,j}^{(s)}$,
future $e$-value pairs $(e_{i,j}^{(s)},\bar e_{i,j}^{(s)})$ for $j\geq t$,
and terminal target coefficients $c_i^{(s)}$. We use the conventions
$R_{i,j}^{(s)}=0$, $c_i^{(s)}=0$, and
$(e_{i,j}^{(s)},\bar e_{i,j}^{(s)})=(1,1)$ outside the simulated horizon or
when $i>M^{(s)}$. Define the simulated cumulative increments
\begin{equation}\label{eq:est-simulated-increments}
  G_{i,t}^{(s)}=
  \sum_{j=t}^{N^{(s)}}R_{i,j}^{(s)}(e_{i,j}^{(s)}-1),\qquad
  \bar G_{i,t}^{(s)}=
  \sum_{j=t}^{N^{(s)}}R_{i,j}^{(s)}(\bar e_{i,j}^{(s)}-1).
\end{equation}

For a fixed finite cap $W>0$, we use a constant-weight approximation within
each completion: a candidate current pair $(w_{i,t},\bar w_{i,t})$ is also
used for all simulated future observations of decision $i$. The pair selected
for $i\in\Acal_t$ solves
\begin{equation}\label{eq:pooled_lp}
\begin{aligned}
\min_{\substack{0\leq w_{i,t},\bar w_{i,t}\leq W\\
                 (z_i^{(s)})_{s=1}^{S}\in\RR^S}}
&\quad \frac1S\sum_{s=1}^{S}z_i^{(s)}\\
\text{subject to}\quad
z_i^{(s)}
&\geq c_i^{(s)}-A_{i,t-1}-w_{i,t}G_{i,t}^{(s)},\\
z_i^{(s)}
&\geq-\bar A_{i,t-1}-\bar w_{i,t}\bar G_{i,t}^{(s)},
\qquad s\in[S].
\end{aligned}
\end{equation}
Equation~\eqref{eq:pooled_lp} replaces the conditional expectation in
\eqref{eq:oracle_single_1} by a simulated average, replaces the continuation
policy by a constant-weight approximation, and imposes the finite cap $W$.
It is a linear program. After solving it, only its current pair
$(w_{i,t},\bar w_{i,t})$ is used; the problem is re-solved after the history is
updated.

\begin{algorithm}[t]
\caption{Predictable pooled-LP tuning for estimation}
\label{alg:pooled_lp}
\begin{algorithmic}[1]
\REQUIRE Number of completions $S$, finite cap $W>0$, and a conditional
trajectory simulator.
\STATE Initialize $A_{i,0}=\bar A_{i,0}=0$ for every $i$.
\FOR{$t=1,2,\ldots,N$}
  \STATE Fit or update $\hat Q_t(\cdot\mid\Fcal_{t-1})$ from the observed history.
  \STATE Draw $S$ complete continuations and compute
  $c_i^{(s)}$, $G_{i,t}^{(s)}$, and $\bar G_{i,t}^{(s)}$.
  \FORALL{$i\in\Acal_t$}
    \STATE Set $(w_{i,t},\bar w_{i,t})$ to a minimizer of
    Equation~\eqref{eq:pooled_lp}, using a fixed deterministic tie-breaking rule.
  \ENDFOR
  \STATE Observe the time-$t$ $e$-value pairs.
  \FORALL{$i\in\Acal_t$}
    \STATE Set $A_{i,t}\gets
    A_{i,t-1}+w_{i,t}R_{i,t}(e_{i,t}-1)$.
    \STATE Set
    $\bar A_{i,t}\gets\bar A_{i,t-1}+\bar w_{i,t}R_{i,t}(\bar e_{i,t}-1)$.
  \ENDFOR
\ENDFOR
\RETURN $\displaystyle
  \hat\theta_*^U=\sum_{i=1}^{M}
  \max\{c_i-A_{i,N},-\bar A_{i,N}\}$.
\end{algorithmic}
\end{algorithm}

\subsection{Empirical Tuning for Confidence Bounds}

We use the same conditional simulator to tune the confidence set while
preserving Assumption~\ref{assump:betting_and_transfers}. Fix in advance a
predictable schedule of betting fractions
$(\lambda_{i,t},\bar\lambda_{i,t})\in[0,1]^2$ and tune only the shared-wealth
transfers $b_{i,t}$. Recall from Equation~\eqref{eq:16} that
\begin{equation}\label{eq:ci-pool-definitions}
  B_t=1+\sum_{i=1}^{M}\min\{A_{i,t},\bar A_{i,t}\},\qquad
  D_{i,t}=(A_{i,t}-\bar A_{i,t})_+,\qquad
  \bar D_{i,t}=(\bar A_{i,t}-A_{i,t})_+.
\end{equation}
The current transfers are nonnegative, $\Fcal_{t-1}$-measurable, and satisfy
$\sum_{i\in\Acal_t}b_{i,t}\leq B_{t-1}$.

For a finite confidence set $C$, define its width by
\begin{equation}\label{eq:ci-width-definition}
  \operatorname{wid}(C)=
  \begin{cases}
    \max C-\min C,&C\neq\varnothing,\\
    0,&C=\varnothing.
  \end{cases}
\end{equation}
Let $\mathfrak B_t$ be the class of predictable continuation-transfer policies
$(b_{i,j})_{i\in\Acal_j,\,j\geq t}$ satisfying
Assumption~\ref{assump:betting_and_transfers}, with the betting fractions held
fixed. The exact oracle criterion at time $t$ is the minimum conditional
expected terminal confidence-set width
\begin{equation}\label{eq:ci-oracle}
  \mathcal R_t^{\mathrm{oracle}}
  :=\inf_{\boldsymbol b_{t:N}\in\mathfrak B_t}
  \EE_{\tilde Q}\!\left[
    \operatorname{wid}\{C(\alpha,F_N)\}
    \;\middle|\;\Fcal_{t-1}
  \right].
\end{equation}
Only the current transfers from an oracle minimizer would be implemented, and
the problem would be solved again at time $t+1$. Equation~\eqref{eq:ci-oracle}
is the desired minimum expected-width objective. It is generally nonconvex.
For example, let $M=N=1$, $c_1=1$,
$\lambda_{1,1}=\bar\lambda_{1,1}=1$, and $b_{1,1}=b\in(0,1]$. The label
$v_1=1$ is retained exactly when
$e_{1,1}<1+(\alpha^{-1}-1)/b$, and $v_1=0$ is retained under the analogous
condition for $\bar e_{1,1}$. Hence
\begin{equation}\label{eq:ci-oracle-example}
  \EE_{\tilde Q}[\operatorname{wid}\{C(\alpha,F_N)\}\mid\Fcal_0]
  =\tilde Q\!\left(
    \max\{e_{1,1},\bar e_{1,1}\}
    <1+\frac{\alpha^{-1}-1}{b}
  \right),
\end{equation}
a step function of $b$.

\paragraph{A convex width surrogate.}
Write
\begin{equation}\label{eq:ci-score-definition}
  S_N(v):=K_N(v)-1
  =\sum_{i=1}^{M}\{v_iA_{i,N}+(1-v_i)\bar A_{i,N}\},
  \qquad h_\alpha:=\alpha^{-1}-1.
\end{equation}
By Equation~\eqref{eq:confidence_interval}, $v$ is retained if and only if
$S_N(v)<h_\alpha$. For every $\eta\geq0$, relaxing the retained-label
constraint separately for the upper and lower endpoints gives
\begin{equation}\label{eq:ci-surrogate}
\begin{aligned}
  \operatorname{wid}\{C(\alpha,F_N)\}
  &\leq [\Psi_\eta]_+,\\
  \Psi_\eta
  &:=2\eta h_\alpha+
  \sum_{i=1}^{M}
  \max\!\left\{
    c_i-\eta(A_{i,N}+\bar A_{i,N}),
    -2\eta A_{i,N},
    -2\eta\bar A_{i,N}
  \right\}\\
  &=2\eta(\alpha^{-1}-B_N)+
  \sum_{i=1}^{M}(c_i-\eta\Delta_{i,N})_+,
  \qquad
  \Delta_{i,N}:=|A_{i,N}-\bar A_{i,N}|,
\end{aligned}
\end{equation}
where $[x]_+=\max\{x,0\}$. To see the first inequality, add
$\eta\{h_\alpha-S_N(v)\}>0$ to the objective for each retained $v$, then
maximize over all $v\in\{0,1\}^M$; apply the same argument to the negative
lower endpoint. The maximizations separate coordinatewise. The final identity
uses
$B_N=1+\sum_i\min\{A_{i,N},\bar A_{i,N}\}$ and
$\Delta_{i,N}=\max\{A_{i,N},\bar A_{i,N}\}-
\min\{A_{i,N},\bar A_{i,N}\}$. Thus $\Psi_\eta$ is a convex,
piecewise-linear surrogate that rewards evidence only until it separates the
two labels of decision $i$ by approximately $c_i/\eta$.

\paragraph{A linear program for the current transfers.}
To obtain a tractable one-step problem, we use a rollout policy that makes the
terminal evidence affine in the current transfer. The amount $b_{i,t}$ is
reserved for decision $i$ for the rest of the simulated trajectory, so no
later reallocation occurs across decisions. Within decision $i$, its two
label-specific balances are reinvested using the fixed betting-fraction
schedule. For completion $s$, define the simulated growth factors
\begin{align}
  E_{i,t:N}^{(s)}
  &:=\prod_{j=t}^{N^{(s)}}
  \left\{1+\lambda_{i,j}R_{i,j}^{(s)}(e_{i,j}^{(s)}-1)\right\},\nonumber\\
  \bar E_{i,t:N}^{(s)}
  &:=\prod_{j=t}^{N^{(s)}}
  \left\{1+\bar\lambda_{i,j}R_{i,j}^{(s)}
  (\bar e_{i,j}^{(s)}-1)\right\}.
  \label{eq:ci-growth-factors}
\end{align}
More explicitly, after simulated time $j\geq t$ the two balances are
\begin{equation}\label{eq:ci-rollout-balances}
\begin{aligned}
  H_{i,j}^{(s)}
  &:=(D_{i,t-1}+b_{i,t})
    \prod_{k=t}^{j}
    \{1+\lambda_{i,k}R_{i,k}^{(s)}(e_{i,k}^{(s)}-1)\},\\
  \bar H_{i,j}^{(s)}
  &:=(\bar D_{i,t-1}+b_{i,t})
    \prod_{k=t}^{j}
    \{1+\bar\lambda_{i,k}R_{i,k}^{(s)}
    (\bar e_{i,k}^{(s)}-1)\}.
\end{aligned}
\end{equation}
At each later simulated time $j>t$, this is represented in
Assumption~\ref{assump:betting_and_transfers} by the transfer
$b_{i,j}^{(s)}=R_{i,j}^{(s)}
\min\{H_{i,j-1}^{(s)},\bar H_{i,j-1}^{(s)}\}$. Thus, whenever
$R_{i,j}^{(s)}=1$,
$D_{i,j-1}^{(s)}+b_{i,j}^{(s)}=H_{i,j-1}^{(s)}$ and
$\bar D_{i,j-1}^{(s)}+b_{i,j}^{(s)}=\bar H_{i,j-1}^{(s)}$; when
$R_{i,j}^{(s)}=0$, the balances are simply carried forward. Because the
reserved balances are funded by the time-$t$ budget, these implied future
transfers satisfy the shared-wealth constraint. The resulting terminal
evidence is affine in the current transfer:
\begin{equation}\label{eq:ci-affine}
\begin{aligned}
  A_i^{(s)}(b_{i,t})
  &=A_{i,t-1}+(D_{i,t-1}+b_{i,t})(E_{i,t:N}^{(s)}-1),\\
  \bar A_i^{(s)}(b_{i,t})
  &=\bar A_{i,t-1}+(\bar D_{i,t-1}+b_{i,t})
    (\bar E_{i,t:N}^{(s)}-1).
\end{aligned}
\end{equation}
For a common multiplier $\eta_t\geq0$, the current transfers are selected by
the epigraph linear program
\begin{equation}\label{eq:ci-lp}
\begin{aligned}
\min_{\substack{b_{i,t}\geq0,\;i\in\Acal_t\\
                 q_i^{(s)}\in\RR,\;i\in\Acal_t,\;s\in[S]}}
&\quad \frac1S\sum_{s=1}^{S}\sum_{i\in\Acal_t}q_i^{(s)}\\
\text{subject to}\quad
q_i^{(s)}
&\geq c_i^{(s)}-\eta_t
  \{A_i^{(s)}(b_{i,t})+\bar A_i^{(s)}(b_{i,t})\},\\
q_i^{(s)}
&\geq-2\eta_t A_i^{(s)}(b_{i,t}),\\
q_i^{(s)}
&\geq-2\eta_t\bar A_i^{(s)}(b_{i,t}),
\qquad i\in\Acal_t,\ s\in[S],\\
\sum_{i\in\Acal_t}b_{i,t}
&\leq B_{t-1}.
\end{aligned}
\end{equation}
The constant $2\eta_t h_\alpha$ and terms for decisions outside
$\Acal_t$ are omitted because they do not depend on the current transfers.
Equations~\eqref{eq:ci-affine}--\eqref{eq:ci-lp} show explicitly that the
program is linear in $(b_{i,t},q_i^{(s)})$.

\paragraph{Choice of the surrogate multiplier.}
From Equation~\eqref{eq:ci-surrogate}, $\eta$ assigns decision $i$ the
evidence-separation target $c_i/\eta$. From
Equation~\eqref{eq:ci-pool-definitions}, a fixed label for decision $i$ is
excluded once its excess evidence reaches $\alpha^{-1}-B_N$; further evidence
on that side cannot reduce the confidence-set width. We therefore use the
predictable plug-in choice
\begin{equation}\label{eq:ci-eta}
  \eta_t=\frac{\bar c_t}{\alpha^{-1}-B_{t-1}},\qquad
  \bar c_t=
  \frac{\displaystyle\sum_{s=1}^{S}\sum_{i=1}^{M^{(s)}}(c_i^{(s)})^2}
       {\displaystyle\sum_{s=1}^{S}\sum_{i=1}^{M^{(s)}}c_i^{(s)}}.
\end{equation}
If the denominator defining $\bar c_t$ is zero, we set $\eta_t=0$ and
$b_{i,t}=0$. If $B_{t-1}\geq\alpha^{-1}$, every candidate label vector is
already excluded, so we again set $b_{i,t}=0$. When all nonzero target
coefficients equal $c$, $\bar c_t=c$; in particular,
$\bar c_t=1/M$ for a fixed-$M$ correctness proportion.

\begin{algorithm}[t]
\small
\caption{Predictable LP tuning of wealth transfers}
\label{alg:ci-tuning}
\begin{algorithmic}[1]
\REQUIRE Number of completions $S$, predictable betting fractions
$(\lambda_{i,t},\bar\lambda_{i,t})$, and a conditional trajectory simulator.
\STATE Initialize $A_{i,0}=\bar A_{i,0}=0$ for every $i$, so $B_0=1$.
\FOR{$t=1,2,\ldots,N$}
  \STATE Fit or update $\hat Q_t(\cdot\mid\Fcal_{t-1})$ and draw $S$
  complete continuations.
  \STATE Compute $c_i^{(s)}$, $E_{i,t:N}^{(s)}$, and
  $\bar E_{i,t:N}^{(s)}$ for every completion.
  \IF{$B_{t-1}<\alpha^{-1}$ and
      $\sum_{s=1}^{S}\sum_{i=1}^{M^{(s)}}c_i^{(s)}>0$}
    \STATE Set $\eta_t$ by Equation~\eqref{eq:ci-eta}.
    \STATE Set $(b_{i,t})_{i\in\Acal_t}$ to a minimizer of
    Equation~\eqref{eq:ci-lp}, using a fixed deterministic tie-breaking rule.
  \ELSE
    \STATE Set $b_{i,t}=0$ for every $i\in\Acal_t$.
  \ENDIF
  \FORALL{$i\in\Acal_t$}
    \STATE Set $w_{i,t}=\lambda_{i,t}(D_{i,t-1}+b_{i,t})$ and
    $\bar w_{i,t}=\bar\lambda_{i,t}(\bar D_{i,t-1}+b_{i,t})$.
  \ENDFOR
  \STATE Observe the time-$t$ $e$-value pairs.
  \FORALL{$i\in\Acal_t$}
    \STATE Update
    $A_{i,t}$ and $\bar A_{i,t}$ by Equation~\eqref{eq:est-running}.
  \ENDFOR
  \STATE Update $B_t$, $D_{i,t}$, and $\bar D_{i,t}$ by
  Equation~\eqref{eq:ci-pool-definitions}.
\ENDFOR
\RETURN $C(\alpha,F_N)$ from Equation~\eqref{eq:confidence_interval}.
\end{algorithmic}
\end{algorithm}

At each time, Algorithm~\ref{alg:ci-tuning} chooses nonnegative predictable
transfers satisfying the shared budget and then constructs the weights exactly
as required by Assumption~\ref{assump:betting_and_transfers}. Therefore the
confidence set retains the uniform coverage guarantee of
Theorem~\ref{thm:coverage}; misspecification of the trajectory simulator can
affect width, but not validity.

\newpage

\section{\texttt{NetworkX} Case Study Supplemental Details}
\paragraph{Agents and task.} The agents are Claude Opus 5 (\texttt{claude-opus-5}) and Claude Haiku 4.5 (\texttt{claude-haiku-4-5-20251001}), both at high effort. The code base is \texttt{NetworkX} at upstream commit \texttt{2622da75c}. Each agent ran 50 sessions in sequence. A session received the current source tree and the number of changes kept so far, and was asked to make one localized change that reduces runtime. The prompt required the agent to run unit tests after changing the code to ensure that the edit did not functionally change the behavior of the software.

\paragraph{Trajectory.} Each code-change was applied to the tree seen by all later sessions. Each change is therefore proposed with knowledge of its predecessors, and the changes of one agent are dependent. Both agents completed all 50 sessions before any measurement described below was taken, so the measurements could not influence the changes.

\paragraph{Measurement.} Every change that altered the code was measured against its parent tree. One measurement is a timing pair. A fresh random graph is drawn from a workload family determined by the target the agent declared, using a seed that the agent never saw. The parent tree and the changed tree each run the workload on that graph in a separate process, one immediately after the other, and the order alternates from pair to pair. Each run performs a warm-up and then times a repeat loop. The pair is kept only if the two runs print the same output checksum, and no pair failed this check. The contrast is
\begin{equation*}
    Y_{i,t}=\log T^{\mathrm{new}}_{i,t}-\log T^{\mathrm{old}}_{i,t},
\end{equation*}
where $T^{\mathrm{new}}_{i,t}$ and $T^{\mathrm{old}}_{i,t}$ are the loop times of the changed tree and of the parent tree, so that negative values favor the change. The changes fall into 14 workload families, among them betweenness centrality, clustering, shortest paths, topological generations, and iteration over neighbors and edges. Generation and measurement ran on a single Linux host (AMD EPYC 9R14, 32 virtual CPUs, Python 3.11).

\paragraph{Tuning Procedures}
We generate $S=64$ conditional continuations from a scale-equivariant posterior-predictive model with standardized-effect prior variance $4$. Each simulated continuation reproduced the sampling rule used in the experiment: stopping was checked after 10 pairs and every 5 pairs thereafter, and the trajectory ended when either $E_{i,t}$ or $\bar{E}_{i,t}$ reached $100$ or when $150$ pairs had been observed. The pooled-LP estimator was retuned before each new measurement block: initially before the first 10 pairs, and subsequently before every block of 5 pairs. The selected weights, constrained to lie in $[0,0.05]$, were held fixed over the next block and retuned again if sampling continued. For inference, the shared-wealth procedure was updated after each observed pair, using betting fractions $\lambda_{i,t}=\bar\lambda_{i,t}=1$. Transfers $b_{i,t}$ were selected by minimizing the convex expected-width surrogate over a 33-point grid of the available wealth; thus, our implementation used a discretized approximation to the continuous transfer LP in Algorithm~2. Simulated evidence increments were truncated at $10^6$ for numerical stability when selecting tuning parameters, Truncation was not applied to the observed evidence used to construct the confidence set. 

\subsection{The $e$-Processes of the \texttt{NetworkX} Case Study}\label{app:eprocess}

Fix a change $i$ and a minimum speedup $p$, and suppress both from the notation. Let $X_n=Y_n+\delta_p$ be the shifted contrasts, with mean $\mu$ and variance $\sigma^2$, so that $\theta=1$ if and only if $\mu\le0$. Because $Y_n$ is the logarithm of the ratio of the new runtime to the old and $\delta_p=-\log(1-p/100)$, the condition $\mu\le0$ says that the geometric mean of this ratio is at most $1-p/100$, a relative reduction of at least $p\%$. Let $\phi$ be the standard normal density and fix a scale $s>0$. The pair of processes used in Section \ref{sec:experiments} is $E_0=\bar E_0=1$ and, for $n\ge1$,
\begin{equation}\label{eq:eprocess-integral}
\begin{aligned}
    E_n&=\frac{\displaystyle\int_0^\infty\left\{\int_0^\infty\prod_{k=1}^n\frac1\sigma\,\phi\!\left(\frac{X_k}{\sigma}-\xi\right)\frac{\mathrm{d}\sigma}{\sigma}\right\}\frac2s\,\phi\!\left(\frac{\xi}{s}\right)\mathrm{d}\xi}
              {\displaystyle\int_0^\infty\prod_{k=1}^n\frac1\sigma\,\phi\!\left(\frac{X_k}{\sigma}\right)\frac{\mathrm{d}\sigma}{\sigma}},\\[1ex]
    \bar E_n&=\frac{\displaystyle\int_{-\infty}^0\left\{\int_0^\infty\prod_{k=1}^n\frac1\sigma\,\phi\!\left(\frac{X_k}{\sigma}-\xi\right)\frac{\mathrm{d}\sigma}{\sigma}\right\}\frac2s\,\phi\!\left(\frac{\xi}{s}\right)\mathrm{d}\xi}
              {\displaystyle\int_0^\infty\prod_{k=1}^n\frac1\sigma\,\phi\!\left(\frac{X_k}{\sigma}\right)\frac{\mathrm{d}\sigma}{\sigma}} .
\end{aligned}
\end{equation}
Each numerator is the Gaussian likelihood of the contrasts with mean $\xi\sigma$ and standard deviation $\sigma$, integrated over $\sigma$ against the right Haar measure $\mathrm{d}\sigma/\sigma$ and over the standardized effect $\xi=\mu/\sigma$ against a half-normal distribution with scale $s$, supported on $(0,\infty)$ for $E_n$ and on $(-\infty,0)$ for $\bar E_n$. The denominator is the same likelihood at $\xi=0$, integrated over $\sigma$. Both integrals are available in closed form. Let $S_n=\sum_{k=1}^nX_k$ and $V_n=\sum_{k=1}^nX_k^2$, and define
\begin{equation*}
    \rho_n=\frac{S_n}{\sqrt{V_n}},\qquad a_n=n+\frac1{s^2},\qquad \tau_n=\frac{\rho_n\sqrt n}{\sqrt{a_n-\rho_n^2}} .
\end{equation*}
Then
\begin{equation}\label{eq:eprocess-closed-form}
    E_n=\frac{2}{s\sqrt{a_n}}\left(1-\frac{\rho_n^2}{a_n}\right)^{-n/2}F_n(\tau_n),
    \qquad
    \bar E_n=\frac{2}{s\sqrt{a_n}}\left(1-\frac{\rho_n^2}{a_n}\right)^{-n/2}F_n(-\tau_n),
\end{equation}
where $F_n$ is the distribution function of the Student $t$ distribution with $n$ degrees of freedom. We take $s=1/2$ throughout, and the $e$-value pairs are the increments $E_n/E_{n-1}$ and $\bar E_n/\bar E_{n-1}$.

 If the contrasts are independent and identically distributed as $\Ncal(\mu,\sigma^2)$ with $\mu$ and $\sigma$ unknown, then \eqref{eq:eprocess-integral} is a mixture likelihood ratio for the scale-invariant statistics $X_k/|X_1|$. The law of these statistics depends on $(\mu,\sigma)$ only through $\mu/\sigma$, and integrating the likelihood against the right Haar measure gives their density \citep{lai1976confidence,perezortiz2024estatistics,lindon2025anytime,wang2025anytime}. By Theorem 4 of \citet{grunwald2025supermartingales}, $(E_n)_{n\ge0}$ is a nonnegative supermartingale whenever $\mu\le0$, and $(\bar E_n)_{n\ge0}$ whenever $\mu\ge0$, for every $\sigma>0$ and with respect to the filtration generated by the scale-invariant statistics. The stopping rule of Section \ref{sec:experiments} depends on the data only through $(E_n,\bar E_n)$, so in this case the pairs satisfy Assumption \ref{assump:e_value_constraint} exactly.

Runtime contrasts need not be Gaussian. For independent and identically distributed contrasts with a finite moment of order $q>2$, a strong approximation of their partial sums by those of a Gaussian sequence shows that, at the boundary $\mu=0$, \eqref{eq:eprocess-closed-form} is asymptotically equivalent to the same statistic computed from Gaussian data, which is an exact $e$-process. In this sense $(E_n)_{n\ge0}$ and $(\bar E_n)_{n\ge0}$ are asymptotic $e$-processes \citep{waudbysmith2024timeuniformcentrallimittheory, bibaut2024nearoptimalnonparametricsequentialtests}.




\end{document}